\documentclass[preprint,12pt]{elsarticle}

\usepackage{amsmath,amsfonts}
\usepackage{algorithmic}
\usepackage{array}
\usepackage[tight,footnotesize]{subfigure}
\usepackage{textcomp}
\usepackage{stfloats}
\usepackage{url}
\usepackage{verbatim}
\usepackage{graphicx}
\def\BibTeX{{\rm B\kern-.05em{\sc i\kern-.025em b}\kern-.08em
    T\kern-.1667em\lower.7ex\hbox{E}\kern-.125emX}}
\usepackage{balance}
\usepackage{booktabs}
\usepackage{caption}
\usepackage{makecell}
\usepackage{adjustbox}
\usepackage{wrapfig}
\usepackage[linesnumbered,ruled]{algorithm2e}
\usepackage{amssymb}
\usepackage{lineno}
\usepackage{float}
\usepackage{algorithmic}
\usepackage[tight,footnotesize]{subfigure}
\usepackage{booktabs}
\usepackage{caption}
\usepackage{makecell}
\usepackage{adjustbox}
\usepackage{wrapfig}
\usepackage{xcolor}
\usepackage[linesnumbered,ruled]{algorithm2e}
\usepackage{listings}
\usepackage{makecell}
\usepackage{pifont}
\newcommand{\cmark}{\ding{51}}%
\newcommand{\xmark}{\ding{55}}%

\journal{Journal Name}

\begin{document}
\sloppy
\setlength{\parskip}{0pt}

\begin{frontmatter}




\title{Human–AI Agent Interaction as a Neuroplastic Training Environment}

\author[label1]{Eranga Bandara}
\ead{cmedawer@odu.edu}

\author[label1]{Ross Gore}
\ead{rgore@odu.edu}

\author[label10]{Asanga Gunaratna}
\ead{asanga.gunaratna@complianceoslab.app}

\author[label1]{Ravi Mukkamala}
\ead{rmukkama@odu.edu}

\author[label15]{Nihal Siriwardanagea}
\ead{nihal@gsi2.com}

\author[label18]{Gihan Siriwardanagea}
\ead{gihan.siriwardanagea@stud.lsmu.lt}

\author[label7]{Sachini Rajapakse}
\ead{sachini.rajapakse@iciclelabs.ai}

\author[label7]{Isurunima Kularathna}
\ead{isurunima.kularathna@iciclelabs.ai}

\author[label7]{Pramoda Karunarathna}
\ead{pramoda.karunarathna@iciclelabs.ai}

\author[label7]{Chalani Rajapakse}
\ead{chalani.rajapakse@iciclelabs.ai}

\author[label1]{Sachin Shetty}
\ead{sshetty@odu.edu}

\author[label1]{Christopher K.\ Rhea}
\ead{crhea@odu.edu}

\author[label5]{Ng Wee Keong}
\ead{awkng@ntu.edu.sg}

\author[label6]{Kasun De Zoysa}
\ead{kasun@ucsc.cmb.ac.lk}

\author[label8]{Amin Hass}
\ead{amin.hassanzadeh@accenture.com}

\author[label14]{Shaifali Kaushik}
\ead{Skaus2000@gmail.com}

\author[label17]{Wathsala Herath}
\ead{wathsala.herath@agentsway.ai}

\author[label13]{Preston Samuel}
\ead{preston.l.samuel.mil@health.mil}

\author[label11]{Anita H.\ Clayton}
\ead{AHC8V@uvahealth.org}

\author[label14]{Atmaram Yarlagadda}
\ead{atmaram.yarlagadda.civ@health.mil}


\address[label1]{Old Dominion University, Norfolk, VA, USA}
\address[label10]{AI Motion Labs, Melbourne, Australia}
\address[label5]{Nanyang Technological University, Singapore}
\address[label6]{University of Colombo, Sri Lanka}
\address[label7]{IcicleLabs.AI}
\address[label8]{Accenture Technology Labs, Arlington, VA, USA}
\address[label17]{Agentsway.AI}
\address[label16]{GSI Scandinavia AB}
\address[label18]{Lithuanian University of Health Sciences}
\address[label11]{Department of Psychiatry and Neurobehavioral Sciences, \\ University of Virginia School of Medicine, Charlottesville, VA, USA}
\address[label13]{Blanchfield Army Community Hospital, Fort Campbell, KY, USA}
\address[label14]{McDonald Army Health Center, Newport News, VA, USA}

\begin{abstract}

Interaction with AI agents has become one of the most frequent activities of everyday digital life. Whether a person is conversing with an assistant, working with a coding copilot, or generating images, the interaction follows a common iterative loop: a request is issued, a result returned, the result appraised, and the request revised. We observe that this loop is a high-frequency stream of contact events --- moments at which a returned result meets a person and a conditioned response may fire before deliberate appraisal --- and that everyday agent interaction is therefore an unrecognised neuroplastic training environment. When a result disappoints, reactive patterns of impatience, perfectionism, frustration, self-criticism, and control are repeatedly evoked, and under the principles of activity-dependent synaptic plasticity each uninterrupted cycle deepens the underlying pathway through long-term potentiation. Because the loop trains through sheer repetition, the ordinary use of AI agents may quietly strengthen the very patterns it provokes. We propose that the same training environment can be engaged to the opposite effect. Drawing on an account of conditioned reactive patterns as physical neurone paths --- encoded through repetition and intensity, activated through a pre-cognitive feeling tone that opens a brief regulatory gap --- we develop a framework in which the interaction loop becomes a setting for deliberate change. At the feeling-tone gap, in place of the reactive re-prompt, a person performs behind-the-scenes observation: watching the underlying neural process operate --- the signal arriving, the feeling tone it evokes, and the conditioned path beginning to route --- so that the cascade does not complete and long-term depression weakens the path rather than long-term potentiation strengthening it. We characterise this practice through three layers of observation, and describe two modes of application: a user-guided mode requiring no change to existing tools, and an agent-assisted mode in which an ordinary agent is lightly configured to support observation at the gap. We illustrate the framework through a worked example of generative image prompting --- a fast, visual, emotionally salient loop in which these reactive patterns surface vividly --- showing how a single frustrating session is, behaviourally, nearly identical whether or not it is observed, yet neurologically opposite.

\end{abstract}


\end{frontmatter}

\section{Introduction}
\label{sec:intro}

Interaction with artificial intelligence agents has, within a few years, 
become one of the most frequent activities of everyday digital life. Hundreds 
of millions of people now converse with assistants, delegate tasks to coding 
copilots, and produce images, text, and designs through generative tools 
\cite{brynjolfsson2023generativeai, noy2023experimental}. Despite the 
diversity of these systems, the interactions they support share a common 
structure: an iterative loop in which a person issues a request, receives a 
result, reacts to that result, and revises the request. A prompt is written; 
an output appears; the person judges it --- often instantly, often 
emotionally --- and prompts again. This loop repeats many times within a 
single session and many sessions within a day.

\begin{figure}[htb]
\centering
\includegraphics[width=\textwidth]{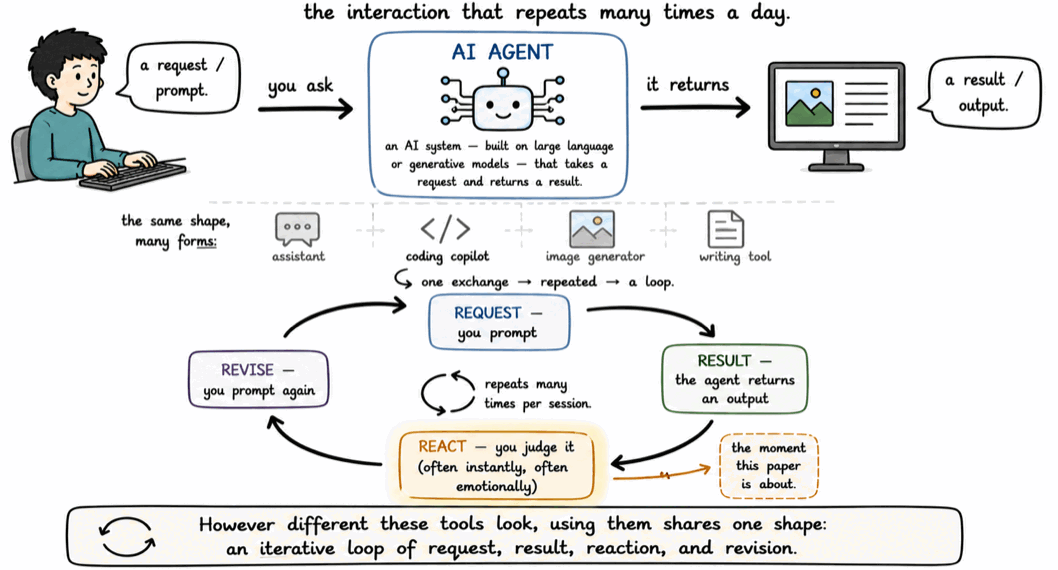}
\caption{What an AI agent is, and why its use forms a loop. A person issues a 
request in natural language; the agent --- an AI system built on a large 
language or generative model --- returns a result, which the person then 
evaluates. Because the result rarely matches the intention on the first 
attempt, the person reacts and issues a revised request, and the exchange 
repeats. However different such tools appear --- conversational assistants, 
coding copilots, image or text generators --- their use shares one shape: an 
iterative loop of request, result, reaction, and revision. The reaction phase 
(highlighted) is the moment this paper concerns.}
\label{fig:whatisagent}
\end{figure}

By an \textit{AI agent} we mean an artificial-intelligence system --- typically 
built on a large language model or a generative model --- that accepts a 
request expressed in natural language and returns a result the user then 
evaluates and acts upon~\cite{agentic-ai, ai-agent-tool-calls}. The category 
spans conversational assistants, coding 
copilots, and generative tools for images, text, and design; what unites them, 
for our purposes, is not their underlying architecture but their mode of use. 
Unlike a static tool that simply executes a fixed command, an agent produces 
an open-ended result that rarely matches the user's intention on the first 
attempt, so the user appraises it and issues a revised request in response. The 
back-and-forth this produces --- request, result, appraisal, revision --- is 
not incidental but intrinsic to how agents are used, and it is this iterative 
loop, rather than any particular model or task, that concerns us 
here~\cite{agent-survey, agentsway} (Figure~\ref{fig:whatisagent}).
 
We make a simple observation about this loop with, we argue, significant 
consequences. Each moment at which a result is received and reacted to is a 
\textit{contact event}: a point at which a stimulus meets a person and a 
conditioned response may be activated before any deliberate appraisal has 
occurred \cite{ledoux1996, ohman2005}. Because the interaction loop is a 
high-frequency stream of such contact events, everyday agent interaction is an 
unrecognised neuroplastic training environment --- a setting that shapes the 
brain through sheer repetition. When a generated image is not 
what was wanted, when a coding agent misunderstands an instruction, or when an 
assistant returns an unhelpful answer, the response that arises is frequently 
not neutral. Impatience, frustration, perfectionism, self-criticism, and the 
urge to control the tool are repeatedly evoked, and each of these is the 
activation of a conditioned reactive pattern. Under the well-established 
principles of activity-dependent synaptic plasticity, every uninterrupted 
repetition of such a pattern strengthens the underlying pathway through 
long-term potentiation \cite{hebb1949, bliss1993, malenka2004, citri2008}. The 
ordinary use of AI agents may therefore be quietly training the very reactive 
patterns it repeatedly provokes.

A training environment, however, shapes in whatever direction it is used, and 
the same repetition admits the opposite outcome. Deeply encoded 
reactive patterns can be attenuated not by managing their consequences 
downstream but by intervening upstream --- at the brief, pre-cognitive window 
between the arising of a feeling tone and the completion of the reactive 
response. We call this window the feeling tone gap, and the practice that 
engages it \textit{behind-the-scenes observation}: rather than completing the 
reaction, the person watches the underlying process operate --- the signal 
arriving, the feeling tone it evokes, and the conditioned path beginning to 
route --- so that the cascade does not complete and long-term depression 
weakens the path rather than long-term potentiation strengthening it. This 
observation can be approached through three graded layers, and applied within 
ordinary human--AI agent interaction itself, requiring no dedicated 
application \cite{mindgap}. Because the interaction loop already 
supplies a dense, naturally occurring stream of contact events, the same 
training environment that conditions reactive patterns is also one in which 
they can be observed and, over repetition, dissolved.

Two features of agent interaction make it particularly well suited to this 
purpose. First, the loop is \textit{frequent and immediate}: the interval 
between stimulus (the result appearing) and response (the reactive re-prompt) 
is short and clearly bounded, so the feeling-tone window is comparatively easy 
to locate~\cite{towards-rai-xai}. Second, the stakes are typically \textit{low}: a disappointing image or an imperfect draft is a safe, repeatable, and inconsequential 
stimulus, which makes ordinary agent use a gentler training context than the 
high-stakes situations in which reactive patterns are usually encountered.

We describe two ways the framework may be applied. In a \textit{user-guided} 
mode, a person who understands the framework practises observation at the 
feeling-tone gap during their own ordinary agent use, without any change to 
the tools they already use. In an \textit{agent-assisted} mode, an existing 
agent is lightly configured --- at the level of a system prompt or an optional 
interaction mode --- to support observation at the gap, without being turned 
into a bespoke clinical system~\cite{deep-psychiatric}. The latter mode connects this proposal to a 
broader question for the designers of agentic systems: whether agents should 
be built merely to satisfy requests as quickly as possible, or whether they 
might also be designed so that the millions of small reactive moments they 
occasion become opportunities for regulation rather than for entrenchment~\cite{nurolense}.

We illustrate the framework throughout with the case of generative image 
prompting. This interaction loop is an especially clear teaching example 
because it is fast, visual, and emotionally salient: a result appears within 
seconds, the mismatch between intention and output is immediately visible, and 
the reaction it provokes --- disappointment, impatience, the compulsion to 
try ``just one more'' prompt --- is vivid and familiar. The patterns that 
surface in this loop are the same conditioned patterns that operate elsewhere 
in a person's life; the prompting session simply makes them observable at high 
frequency and low cost~\cite{mindgap}.

This paper makes three contributions. First, we characterise the iterative 
human--AI interaction loop as a contact-event cycle, and argue that everyday 
agent interaction is a previously unexamined neuroplastic training environment 
that can entrench or attenuate reactive patterns depending on how the loop is 
engaged. Second, we present a framework for upstream neuroplastic intervention 
that operates within ordinary agent interaction rather than through a 
dedicated application, in both user-guided and agent-assisted modes. Third, we 
develop a worked example in generative image prompting that makes the 
mechanism concrete and yields design implications for the builders of agentic 
systems.

The remainder of this paper is organised as follows. 
Section~\ref{sec:related} reviews related work on affect and frustration in 
human--AI interaction, the neuroscience of plasticity, and observation-based 
approaches to emotion regulation. Section~\ref{sec:theory} sets out the 
theoretical foundations using everyday examples --- the reactive-pattern 
model, the feeling-tone gap, and the opposing mechanisms of long-term 
potentiation and depression --- without reference to AI agents. 
Section~\ref{sec:agents} then shows how these same mechanisms operate in 
human--AI agent environments, characterising the interaction loop as a 
contact-event cycle. Section~\ref{sec:rewiring} presents the framework for 
rewiring through agent interaction and its two modes. 
Section~\ref{sec:example} develops the generative-prompting worked example. 
Section~\ref{sec:discussion} discusses design implications, positioning, and 
limitations, and Section~\ref{sec:conclusion} concludes.

\section{Related Work}
\label{sec:related}

This work draws together three literatures that have not previously been 
connected: the study of affect in human--computer and human--AI interaction, 
the neuroscience of experience-dependent plasticity, and observation-based 
approaches to emotion regulation. We review each in turn and then identify the 
gap the present framework addresses.

\subsection{Affect and Frustration in Human--AI Interaction}
\label{subsec:rw_hci}

That people respond to computers emotionally, and even socially, is a 
long-established finding. Work on the ``media equation'' demonstrated that 
users treat computers as social actors, applying interpersonal expectations 
and emotional responses to them despite knowing they are machines 
\cite{reeves1996media}. The field of affective computing developed from the 
recognition that emotion is central to human interaction with technology and 
that systems might productively sense and respond to it \cite{picard1997affective}. 
Frustration in particular has been studied as a common and consequential 
affective state in computer use: systems have been designed to detect and 
actively defuse user frustration, on the premise that it degrades both 
experience and performance \cite{klein2002frustration}. More recently, as 
interaction with AI systems has become widespread, guidelines for human--AI 
interaction have codified design principles intended to reduce friction and 
manage user expectations across the interaction \cite{amershi2019guidelines}.

Across this literature, however, affect is treated as a state to be managed in 
the service of the task --- friction to be minimised, frustration to be 
detected and defused, expectations to be smoothed. The emotional response is 
understood as a transient reaction with consequences for usability and 
satisfaction. What this literature does not consider is that the affective 
responses evoked during interaction might have consequences beyond the 
interaction itself --- that the repeated firing of a reactive pattern during 
tool use could, through ordinary synaptic plasticity, leave a lasting trace in 
the user. The present work takes exactly this step: it treats the affect 
evoked during agent interaction not merely as a UX variable but as an instance 
of neuroplastic conditioning.

\subsection{Neuroplasticity and Experience-Dependent Change}
\label{subsec:rw_neuro}

The mechanistic basis of the framework is the well-established neuroscience of 
activity-dependent synaptic plasticity. The principle that co-active neurons 
strengthen their connections \cite{hebb1949}, realised in the mechanisms of 
long-term potentiation and long-term depression \cite{bliss1993, malenka2004, 
citri2008, luscher2012}, provides the account of how repeated experience 
physically reshapes neural pathways. That such reshaping follows sustained 
behavioural practice is documented by studies of experience-dependent 
structural change, from the hippocampal changes accompanying spatial expertise 
\cite{maguire2000navigation} to grey-matter changes following motor skill 
acquisition \cite{draganski2004changes}. This literature establishes the 
substrate on which the present framework rests; our contribution is not to the 
neuroscience itself but to identifying a novel, ubiquitous, and previously 
unexamined setting --- everyday agent interaction --- in which these 
mechanisms are routinely engaged.

\subsection{Observation-Based Emotion Regulation}
\label{subsec:rw_observation}

The practice at the centre of the framework belongs to a family of 
observation-based approaches to emotion regulation with substantial clinical 
support. Mindfulness-based cognitive therapy cultivates a decentred 
relationship to thoughts and feelings, treating them as passing mental events 
rather than facts, and is associated with reduced relapse in recurrent 
depression \cite{teasdale1999, segal2002, fresco2007}. Metacognitive therapy 
directs attention to the process of thinking rather than its content 
\cite{wells2009}. The broader emotion-regulation literature characterises the 
top-down, prefrontally mediated modulation of affective responses and its role 
in psychological health \cite{gross1998, ochsner2005, gross2015, aldao2010}. 
That sustained contemplative practice of this kind produces measurable 
structural and functional change in the brain is itself documented 
\cite{holzel2011mindfulness, davidson2003alterations}.

These practices are, however, almost always cultivated in dedicated settings 
--- formal meditation, therapy sessions, structured daily exercises --- and 
their benefit in everyday life depends on transferring a capacity built in 
those settings to the moments where reactive patterns actually fire. The 
present framework differs not in the nature of the observation it proposes but 
in its setting: it locates the practice within an activity people already 
perform many times a day, at the exact moments reactive patterns are triggered, 
so that the reactive moment and the opportunity to observe it coincide.

\subsection{Positioning}
\label{subsec:rw_positioning}

The contribution of this paper lies in the connection of these three 
literatures, which have developed independently. The HCI literature 
establishes that agent interaction reliably evokes affect but treats that 
affect as transient UX friction; the neuroscience literature establishes that 
repeated affective responses physically reshape neural pathways but has not 
examined agent interaction as a setting in which this occurs; and the 
emotion-regulation literature establishes that observation can intercept 
reactive responses but situates the practice outside the moments of ordinary 
technology use. Placing these together yields the paper's central claim: that 
the human--AI interaction loop is a high-frequency, naturally occurring site of 
neuroplastic conditioning, at which the same observation-based mechanism 
studied in clinical settings can be practised --- for good or ill depending on 
how the loop is engaged. To our knowledge, no prior work has characterised 
everyday agent interaction in these terms, nor proposed the interaction loop 
itself as a site for upstream neuroplastic intervention.

Table~\ref{tab:relatedwork} summarises this positioning across the dimensions 
that define the present framework. Each strand of prior work supports some of 
these dimensions but none combines them; in particular, no prior work treats 
the interaction loop itself as the intervention site or locates the practice 
within ordinary, tool-native use.

\begin{table*}[!htb]
\centering
\caption{Comparison with representative related work across the dimensions of 
the proposed framework. \cmark~=~supported; \xmark~=~not supported; 
$\sim$~=~partially supported.}
\label{tab:relatedwork}
\begin{adjustbox}{width=1\textwidth}
\begin{tabular}{lcccccc}
\toprule
\thead{Work} & \thead{Human--AI\\interaction\\focus} & 
\thead{Interaction\\loop as\\intervention site} & 
\thead{Neuroplastic\\framing\\(LTP/LTD)} & 
\thead{Feeling-tone\\gap\\(upstream)} & 
\thead{Behind-the-scenes\\observation /\\three layers} & 
\thead{Everyday /\\no-app\\setting} \\
\midrule
Reeves \& Nass~\cite{reeves1996media}          & \cmark & \xmark & \xmark & \xmark & \xmark & \cmark \\
Picard~\cite{picard1997affective}              & \cmark & \xmark & \xmark & \xmark & \xmark & $\sim$ \\
Klein et al.~\cite{klein2002frustration}       & \cmark & \xmark & \xmark & \xmark & \xmark & $\sim$ \\
Amershi et al.~\cite{amershi2019guidelines}    & \cmark & \xmark & \xmark & \xmark & \xmark & \cmark \\
Draganski et al.~\cite{draganski2004changes}   & \xmark & \xmark & \cmark & \xmark & \xmark & \xmark \\
Teasdale et al.~\cite{teasdale1999}            & \xmark & \xmark & $\sim$ & $\sim$ & $\sim$ & \xmark \\
Segal et al.~\cite{segal2002}                  & \xmark & \xmark & $\sim$ & $\sim$ & $\sim$ & \xmark \\
Wells~\cite{wells2009}                         & \xmark & \xmark & \xmark & $\sim$ & $\sim$ & \xmark \\
Gross~\cite{gross2015}                         & \xmark & \xmark & $\sim$ & \cmark & \xmark & \xmark \\
H\"olzel et al.~\cite{holzel2011mindfulness}   & \xmark & \xmark & \cmark & $\sim$ & $\sim$ & \xmark \\
\midrule
\textbf{This work (proposed)}                  & \cmark & \cmark & \cmark & \cmark & \cmark & \cmark \\
\bottomrule
\end{tabular}
\end{adjustbox}
\end{table*}

\section{Theoretical Foundations}
\label{sec:theory}

This section develops the end-to-end account of conditioned reactive patterns 
that the remainder of the paper builds on. It is deliberately presented as 
background, using everyday real-world examples rather than the human--AI 
interaction setting that is the subject of this work; that setting is taken up 
in Section~\ref{sec:agents}, once the mechanism is in place. We proceed from 
the smallest unit --- the single neurone and its synapse --- to the neurone 
path, to the deeply encoded paths that constitute implicit beliefs, to the 
millisecond-scale affective routing that activates them below awareness, and 
finally to the opposing synaptic mechanisms of long-term potentiation and 
long-term depression that respectively deepen and dissolve them. The argument 
is cumulative: each subsection supplies a component that the later sections 
require.

\subsection{The Neurone Path: The Physical Substrate of Psychological 
Experience}
\label{subsec:neuronepath}

\begin{figure}[H]
\centering
\includegraphics[width=0.98\textwidth]{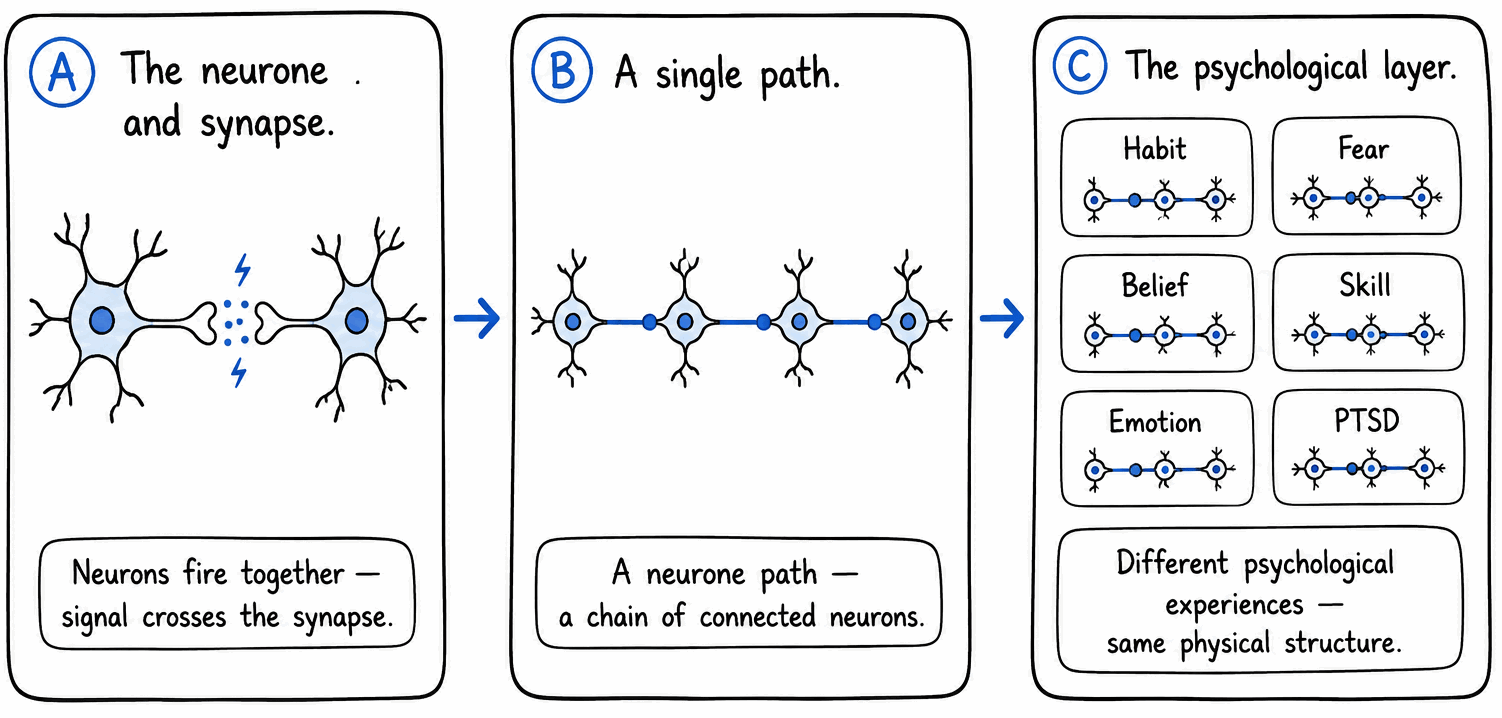}
\caption{What neurone paths are. A single neurone communicates with the next 
by releasing neurotransmitters across the synapse (left). A chain of such 
connected neurons forms a neurone path (centre). Psychological phenomena as 
diverse as habits, fears, beliefs, skills, emotions, and post-traumatic 
stress share the same underlying physical structure --- they differ in 
content and encoding depth, not in kind (right).}
\label{fig:neuronepaths}
\end{figure}

The foundational claim is that psychological experience is physical. A 
thought, a habit, a fear, a belief, or a skill is not an abstract entity 
residing apart from the brain; it is the activity of a specific configuration 
of connected neurons \cite{hebb1949}. Two neurons that fire in close temporal 
succession strengthen the synaptic connection between them, so that subsequent 
activation of the first more readily activates the second \cite{bliss1993, 
malenka2004}. Repeated across chains of neurons, this process produces a 
\textit{neurone path}: a physical sequence of connected neurons that, once 
activated at its origin, propagates activation along its length to a 
characteristic outcome (Figure~\ref{fig:neuronepaths}).

\begin{figure}[H]
\centering
\includegraphics[width=0.98\textwidth]{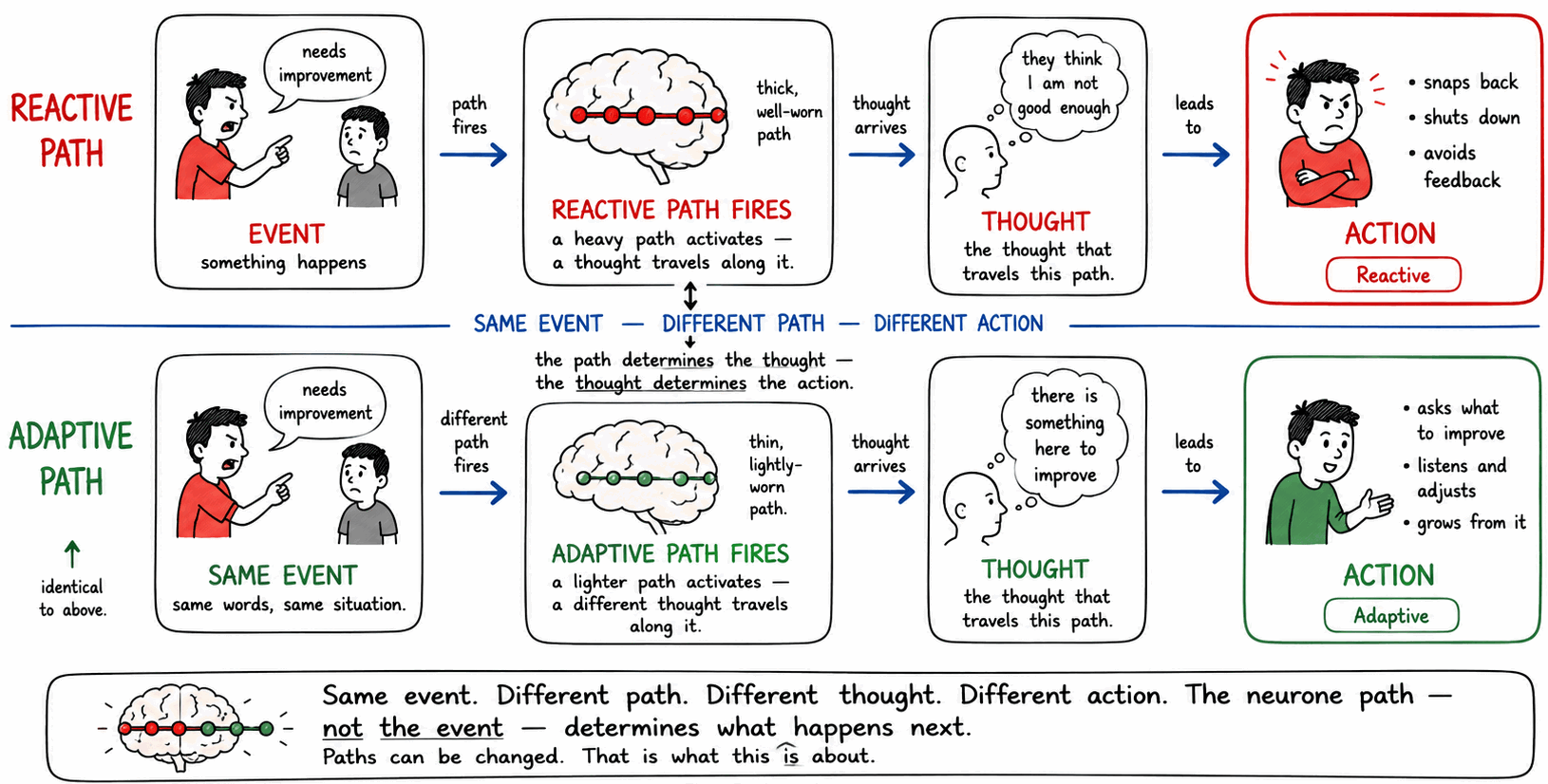}
\caption{How a neurone path works, from event to action. The same event can 
activate different paths in different individuals, and the activated path --- 
not the event itself --- determines the resulting thought and action. A 
heavily encoded path (top) routes rapidly to a reactive outcome; a lighter 
path (bottom) routes to a considered response. The path determines the 
thought; the thought determines the action.}
\label{fig:eventtoaction}
\end{figure}

The psychological significance of a neurone path lies in what it does when it 
fires. A path activated by a particular situation routes toward a 
characteristic experience and behaviour. Critically, this routing occurs 
whether or not the person chooses it: once the path is activated at its 
origin, propagation is automatic. The subjective experience --- the thought, 
the feeling, the urge --- is the activation of the path, not a separate event 
that follows it (Figure~\ref{fig:eventtoaction}).

This reframing has a direct practical consequence. If a reactive pattern is a 
physical path, then changing the pattern is a matter of changing the path --- 
not of willpower, insight alone, or the suppression of an unwanted output. 
The remainder of this section establishes how paths form, how they deepen, 
and how they can be dissolved.

\subsection{How Neurone Paths Form}
\label{subsec:formation}

\begin{figure}[H]
\centering
\includegraphics[width=0.98\textwidth]{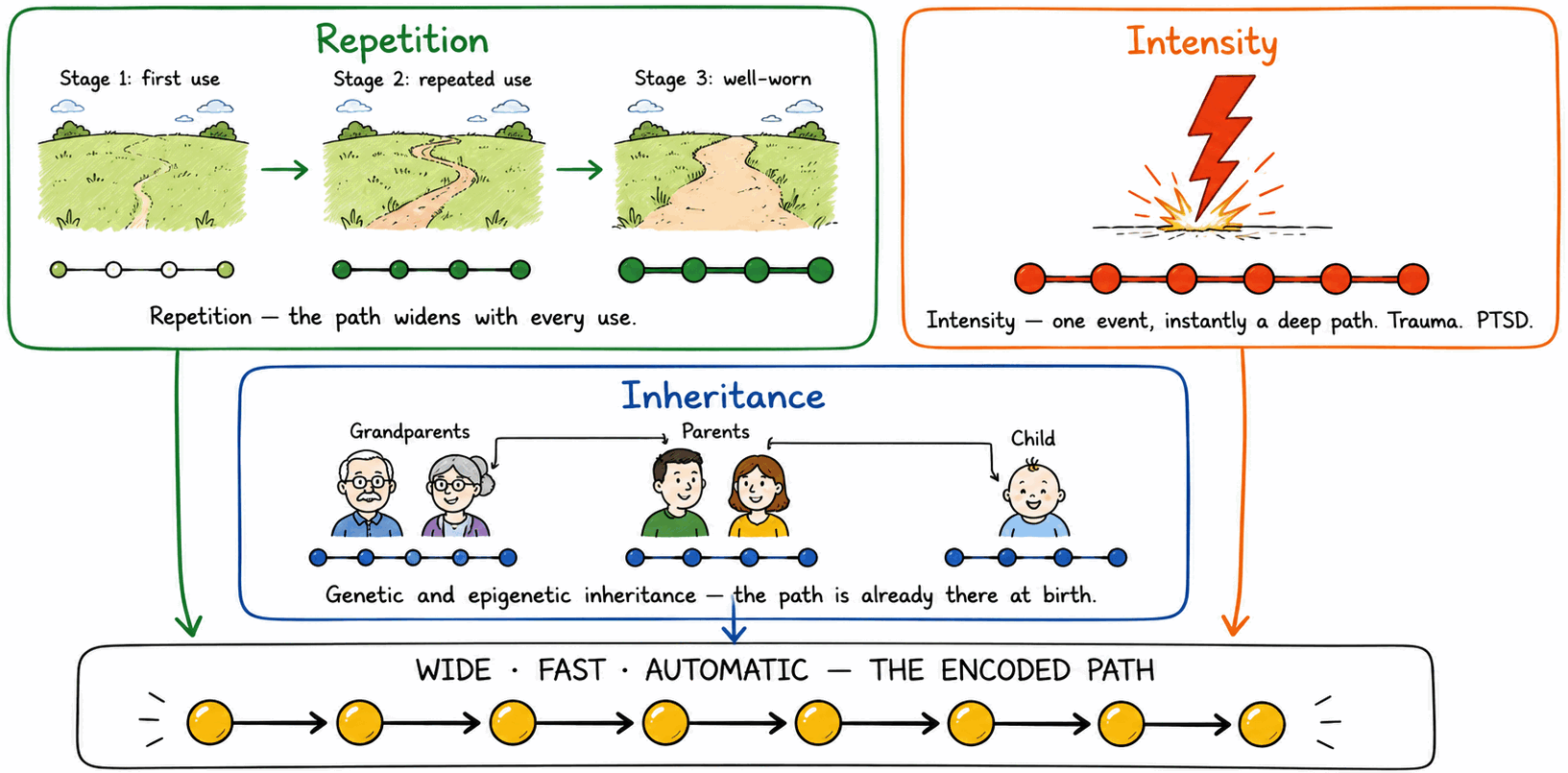}
\caption{How neurone paths form. Three mechanisms encode paths into the 
brain's physical architecture: repetition (many small increments of synaptic 
strengthening over time), single-event intensity (one extreme event producing 
maximal encoding through stress-hormone-accelerated consolidation), and 
genetic or epigenetic inheritance (transmitted predispositions that bias path 
formation).}
\label{fig:pathformation}
\end{figure}

Neurone paths are encoded through three mechanisms. The first is 
\textit{repetition}: a situation and a response that co-occur repeatedly 
strengthen their connection incrementally, each repetition adding a small 
increment of synaptic efficacy \cite{hebb1949, bliss1993}. A child told 
repeatedly that their efforts are not good enough encodes an inadequacy path 
gradually, over years, until it activates automatically in adulthood whenever 
their work is judged --- no single episode is decisive, but each instance adds 
a small increment until the path fires on its own. The second is 
\textit{single-event intensity}: under conditions of extreme arousal, stress 
hormones accelerate synaptic consolidation such that a single event can 
produce the encoding that would otherwise require thousands of repetitions 
\cite{pitman2012, mcnally2003}. The nurse who experiences one overwhelming 
mass-casualty night encodes a danger path in a single exposure. The third is 
\textit{genetic and epigenetic inheritance}: predispositions and the 
consequences of ancestral experience can be transmitted, biasing the ease 
with which particular paths form \cite{roth2009} (Figure~\ref{fig:pathformation}).

These mechanisms are not mutually exclusive; a given path may be inherited as 
a predisposition, established through repetition, and then intensified by a 
single event. What matters is that all three produce the same kind of physical 
structure, and all three are subject to the same mechanisms of deepening and 
dissolution described below.

\subsection{Not All Paths Are Problems}
\label{subsec:notallpaths}

\begin{figure}[H]
\centering
\includegraphics[width=0.98\textwidth]{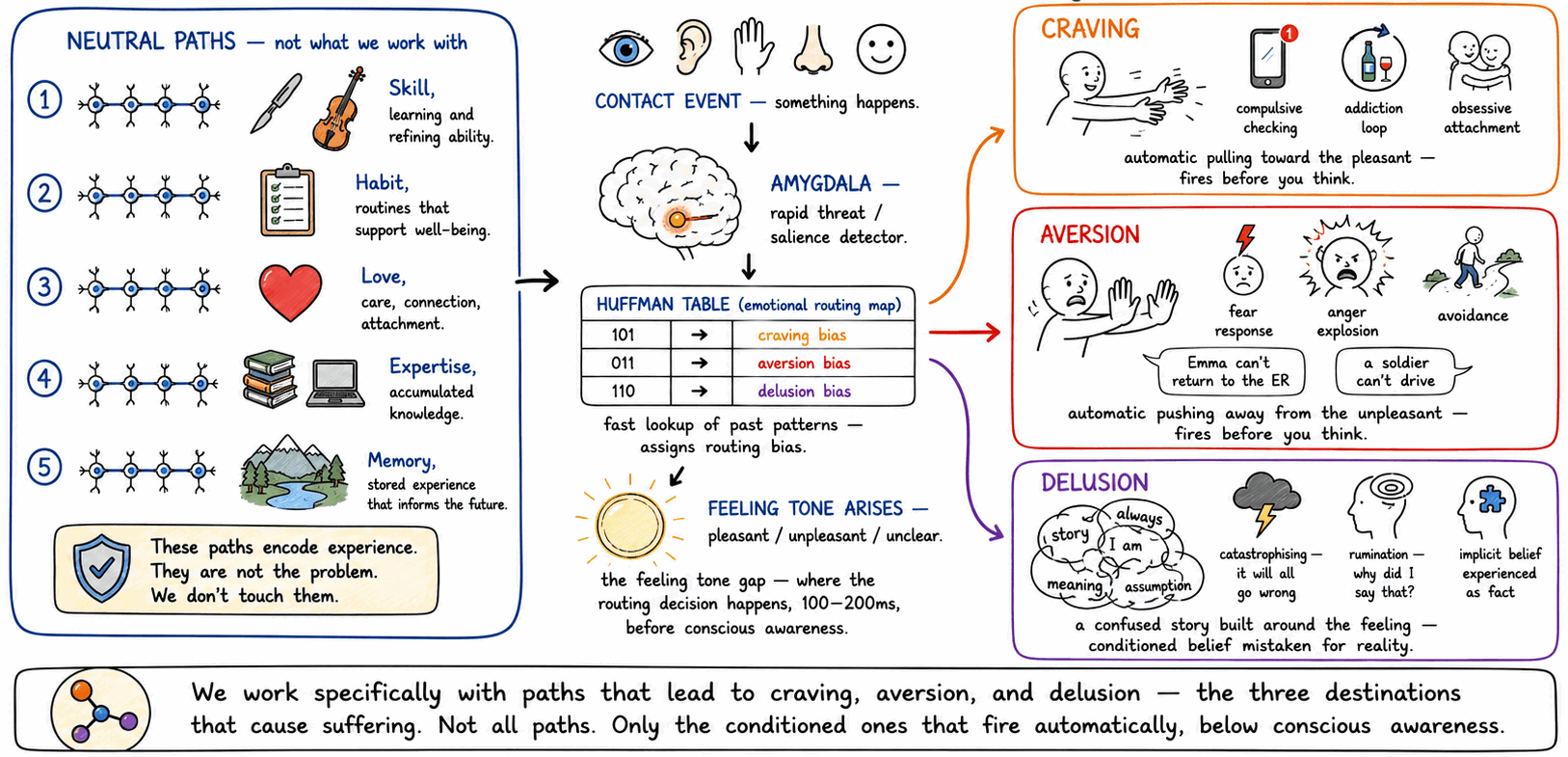}
\caption{Not all paths are problems. The same physical mechanism of encoding 
underlies adaptive paths (skills, knowledge, recognition) and maladaptive 
reactive paths. Only paths that route automatically to reactive suffering --- 
disproportionate, involuntary, and unwanted responses --- are targets for 
intervention. Adaptive encoding is preserved; only the specific reactive paths 
that generate suffering are targeted.}
\label{fig:notallpaths}
\end{figure}

Neurone paths are not pathological in general. The overwhelming majority of 
encoded paths are adaptive or neutral: the skill of reading, the ability to 
ride a bicycle, the recognition of a loved one's face, and the capacity for 
language are all deeply encoded paths, and their automaticity is precisely 
what makes them useful. Deep encoding is not the problem; it is the basis of 
all learning and competence (Figure~\ref{fig:notallpaths}).

The target of intervention is a specific subset: paths that route 
automatically to reactive suffering --- responses that are disproportionate 
to the situation, involuntary, and unwanted by the person who experiences 
them. This scoping matters both practically and ethically: the aim is not 
to flatten affect or erase memory, but to dissolve the specific paths that 
produce reactions the person themselves wishes to be free of.

\subsection{What Is a Situation? Situations Reach Us Through the Senses}
\label{subsec:situation}

\begin{figure}[H]
\centering
\includegraphics[width=0.98\textwidth]{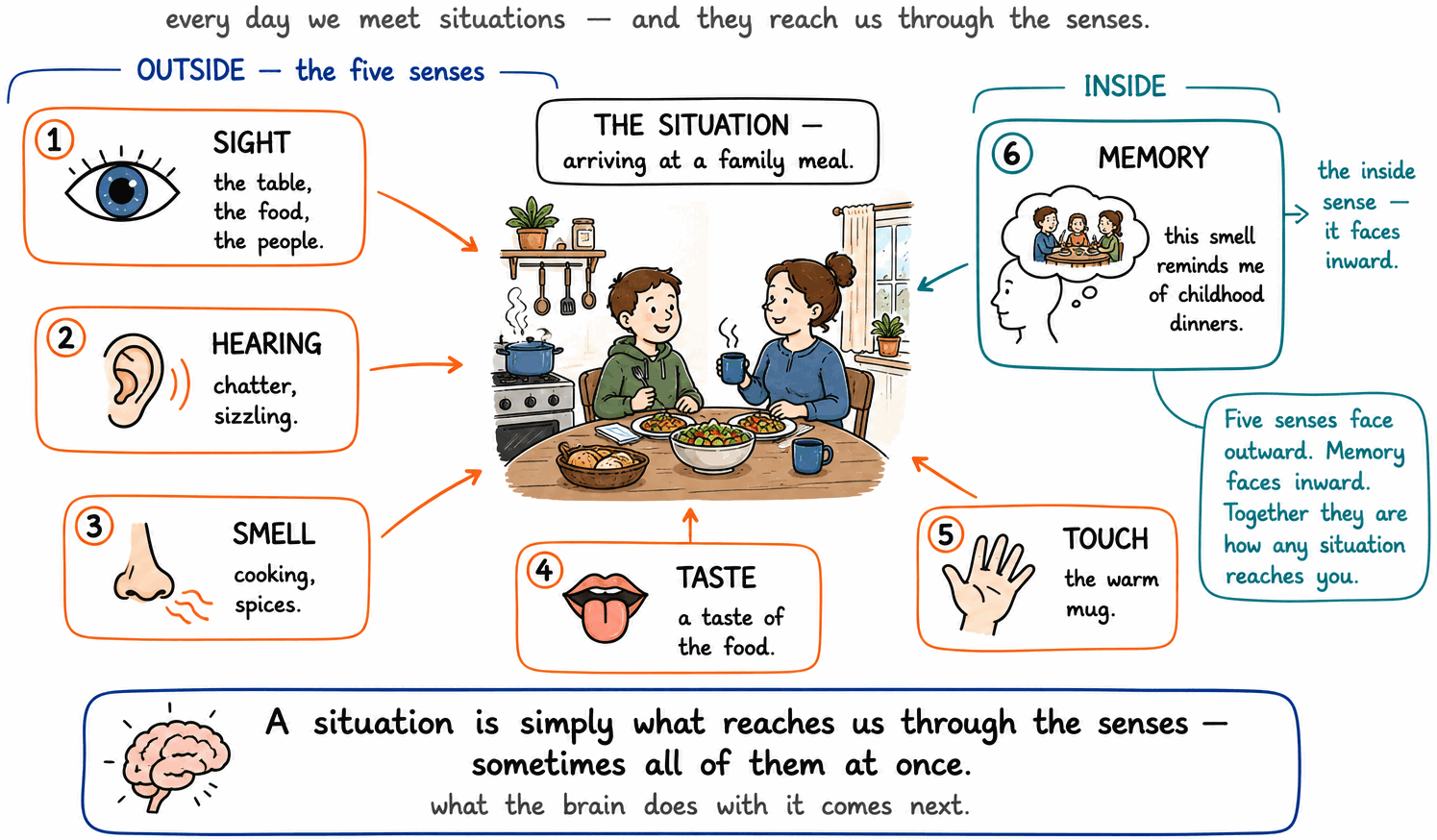}
\caption{What a situation is. An everyday situation --- here, arriving at a 
family meal --- reaches the brain through a closed set of sensory channels: 
the five exteroceptive senses (sight, hearing, smell, taste, touch) and the 
internal channel of memory, which faces inward and can supply content with no 
external input at all. A situation is constituted by whichever channels are 
active, integrated together into a single recognised event.}
\label{fig:whatissituation}
\end{figure}

Before describing how situations are encoded and matched, we must be precise 
about what a situation is. We propose that a situation is not an abstract 
whole but the specific set of sensory inputs through which an event reaches 
the brain. These inputs arrive through a closed set of channels: five 
\textit{exteroceptive} channels that register the external world --- sight, 
hearing, smell, taste, and touch --- and at least one \textit{internal} 
channel, memory, which registers self-generated content such as a surfacing 
recollection or a recalled voice. A felt inner state, such as the sense of 
being alone or unsupported, functions as a further internal channel. This 
closed-set formulation is deliberate: it converts an otherwise arbitrary list 
of situational features into a principled enumeration of the channels through 
which any situation can be registered at all (Figure~\ref{fig:whatissituation}).

A clarification on terminology is warranted here. The brain does not index or 
retrieve whole ``situations'' as discrete objects; mechanistically, what 
propagates and is matched is a \textit{pattern of sensory signals}, appraised 
rapidly and approximately by the amygdala and associated circuits against 
encoded patterns \cite{ledoux1996, ohman2005}. We use ``situation'' throughout 
as a deliberate simplifying abstraction --- a convenient name for that pattern 
of co-occurring sensory signals --- because it makes the account legible 
without altering its mechanistic content. When we speak of a situation being 
matched to a path, we mean that the pattern of sensory signals characterising 
that situation is what the matching operates on.

Two consequences follow. First, a situation is composed only of the channels 
that are actually active; different situations recruit different subsets, and 
no situation requires all channels at once. Second, because memory is a 
genuine internal channel, a situation can be assembled partly or wholly from 
internally generated content --- which is why a recollection alone, with the 
external world entirely quiet, can constitute a situation capable of 
activating a path. This internal channel is central to the account of 
generalisation developed below. Our formulation here is a 
sensory-channel model grounded in multimodal integration; it is worth noting, 
as a historical antecedent rather than a foundation, that classical Buddhist 
analysis likewise treated the mind as a sixth sense base --- alongside the 
five physical senses --- whose objects are mental phenomena including memories 
\cite{bodhi2000abhidhamma, gethin1998foundations}. We do not develop this 
convergence here.

\subsection{A Situation Links to a Neurone Path}
\label{subsec:situationpath}

\begin{figure}[H]
\centering
\includegraphics[width=0.98\textwidth]{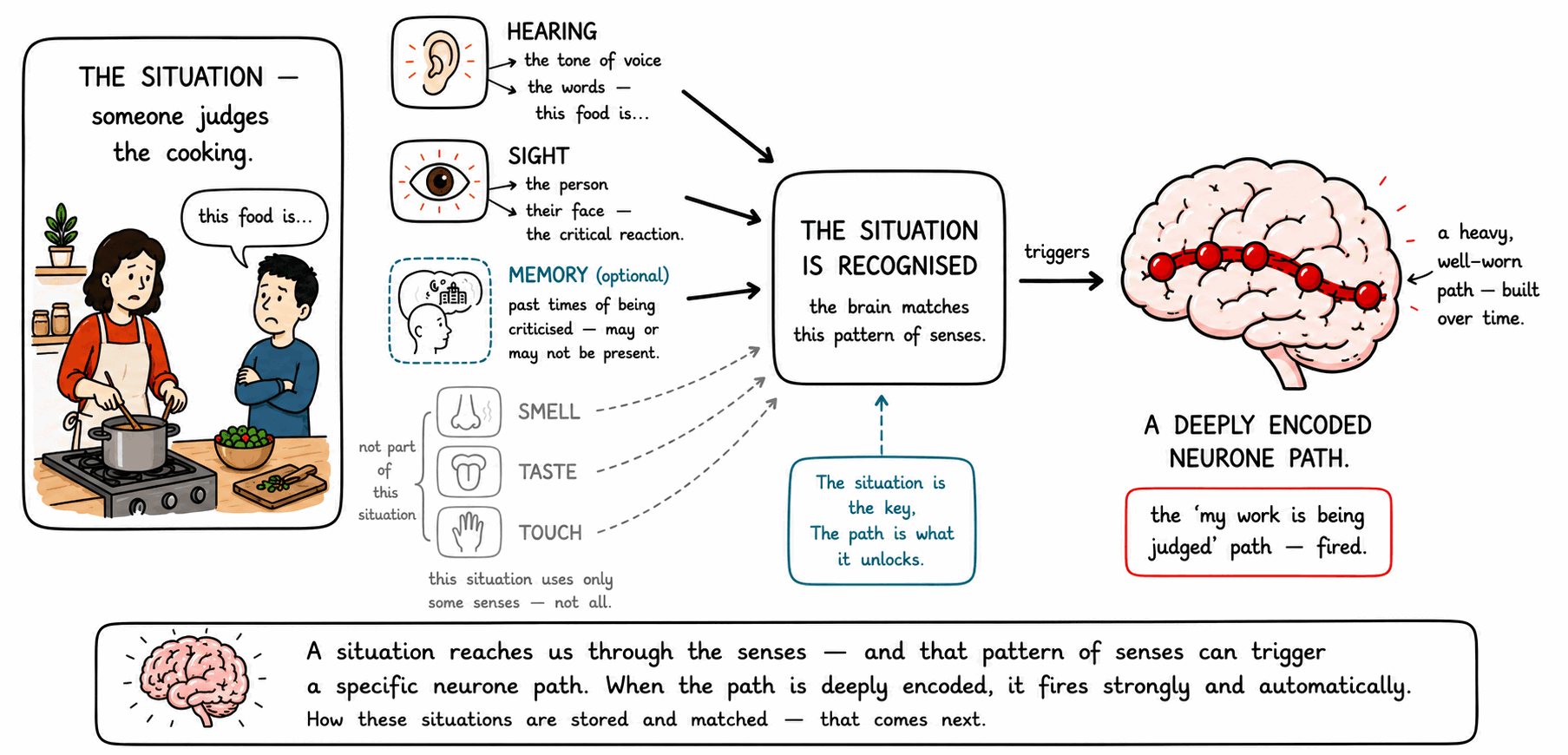}
\caption{A situation links to a neurone path. A specific situation --- someone 
commenting on a cook's food --- reaches the brain through only the channels 
that carry it: hearing (the tone and the words) and sight (the person and 
their facial reaction), with memory as an optional internal contributor. The 
other senses are not part of this situation. This pattern of active channels 
matches, and thereby fires, a deeply encoded neurone path --- here, the 
implicit-belief path ``my work is being judged.''}
\label{fig:situationpath}
\end{figure}

A situation, once registered through its active channels, is matched against 
the brain's store of encoded paths. When the pattern of sensory inputs --- 
appraised rapidly by the amygdala --- matches the signature of an encoded 
path, that path is activated. The situation is thus the \textit{key}, and the 
neurone path is what the key unlocks. Crucially, it is the pattern of sensory 
channels --- not the event in the abstract --- that performs the matching, and 
it is the activated path, not the event, that determines the resulting 
experience and behaviour (Figure~\ref{fig:situationpath}).

This figure also illustrates the point made above: only some channels carry a 
given situation. The cook's situation is composed of hearing and sight; smell, 
taste, and touch play no part in it, even though food is present. The 
situational key is therefore composed of only the channels that carried the 
path's formative attributes --- a subset of the closed channel set, not the 
whole set. This subset structure is what makes the encoding model that 
follows both precise and tractable.

\subsection{Deeply Encoded Paths as Implicit Beliefs: A Situation-Keyed 
Encoding Model}
\label{subsec:huffman}

The most deeply encoded reactive paths function as \textit{implicit beliefs}: 
automatic routing assumptions that determine how a situation is interpreted 
before any deliberate reasoning occurs. A person does not consciously decide 
that criticism means they are inadequate; the felt sense of inadequacy 
arrives first, in milliseconds, and conscious reasoning proceeds from it 
\cite{zajonc1980, barrett2017}. The implicit belief is not a proposition the 
person holds; it is the path that fires.

\begin{figure}[H]
\centering
\includegraphics[width=0.98\textwidth]{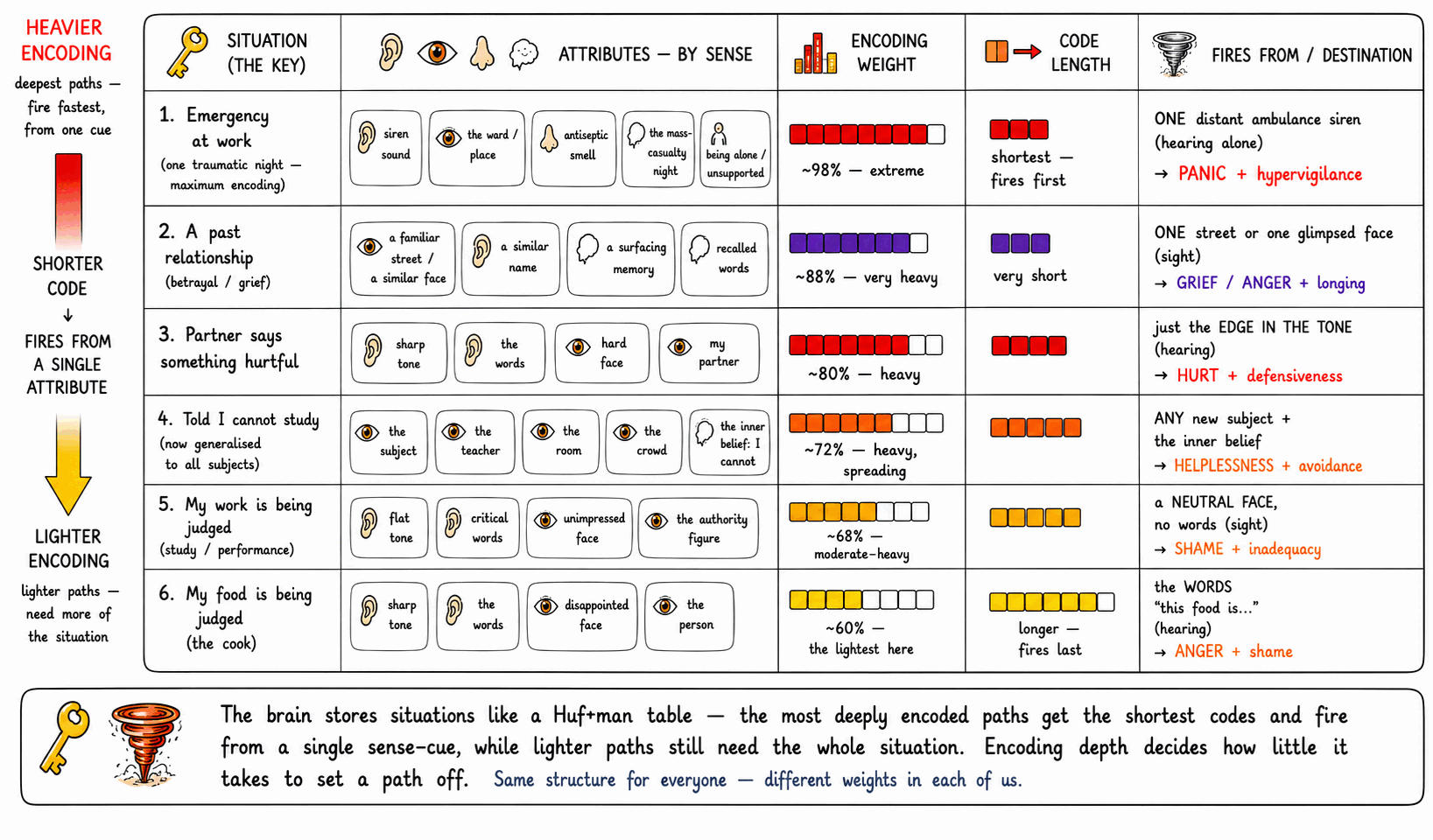}
\caption{The situation-keyed encoding model. Each row represents a conditioned 
reactive path indexed by a situational key comprising attributes such as tone, 
facial expression, words, location, sound, and memory. Deeply encoded paths 
(top rows) carry the greatest weight and fire fastest and most automatically, 
analogous to the shortest codes in Huffman encoding. Incoming situations are 
matched approximately against these keys to generate a pre-cognitive affective 
routing decision in under 200 milliseconds.}
\label{fig:huffmantable}
\end{figure}

We propose that this routing operates as a situation-keyed compressed encoding 
system, analogous to Huffman encoding in information theory \cite{huffman1952}. 
In Huffman encoding, the most frequently occurring symbols are assigned the 
shortest codes, minimising the average length of the encoded message. In the 
brain's affective architecture, the most frequently or intensely encoded 
paths become the shortest, fastest, most automatic routes: they fire first, 
at the lowest stimulus intensity, and below conscious awareness. The 
\textit{key} to this lookup is the situation, decomposed --- as established in 
Section~\ref{subsec:situation} --- into sense-tagged attributes across the 
closed channel set: what is heard (tone, words), seen (face, person, place), 
smelled, and internally registered (memory, inner state). The amygdala 
performs approximate pattern matching of these incoming attributes against 
encoded path signatures, and returns a pre-cognitive affective routing 
decision --- a feeling tone --- in under 200 milliseconds \cite{ledoux1996, 
ohman2005} (Figure~\ref{fig:huffmantable}).

This model extends existing accounts of rapid affective appraisal 
\cite{ledoux1996, ohman2005, barrett2017} by specifying the structure of the 
situational key and the compression principle that determines which paths 
dominate. It also sets up the central phenomenon addressed next: the 
behaviour of these keys as encoding depth increases.

\subsection{Encoding-Depth-Dependent Stimulus Generalisation: The Tornado 
Effect}
\label{subsec:tornado}

\begin{figure}[H]
\centering
\includegraphics[width=0.98\textwidth]{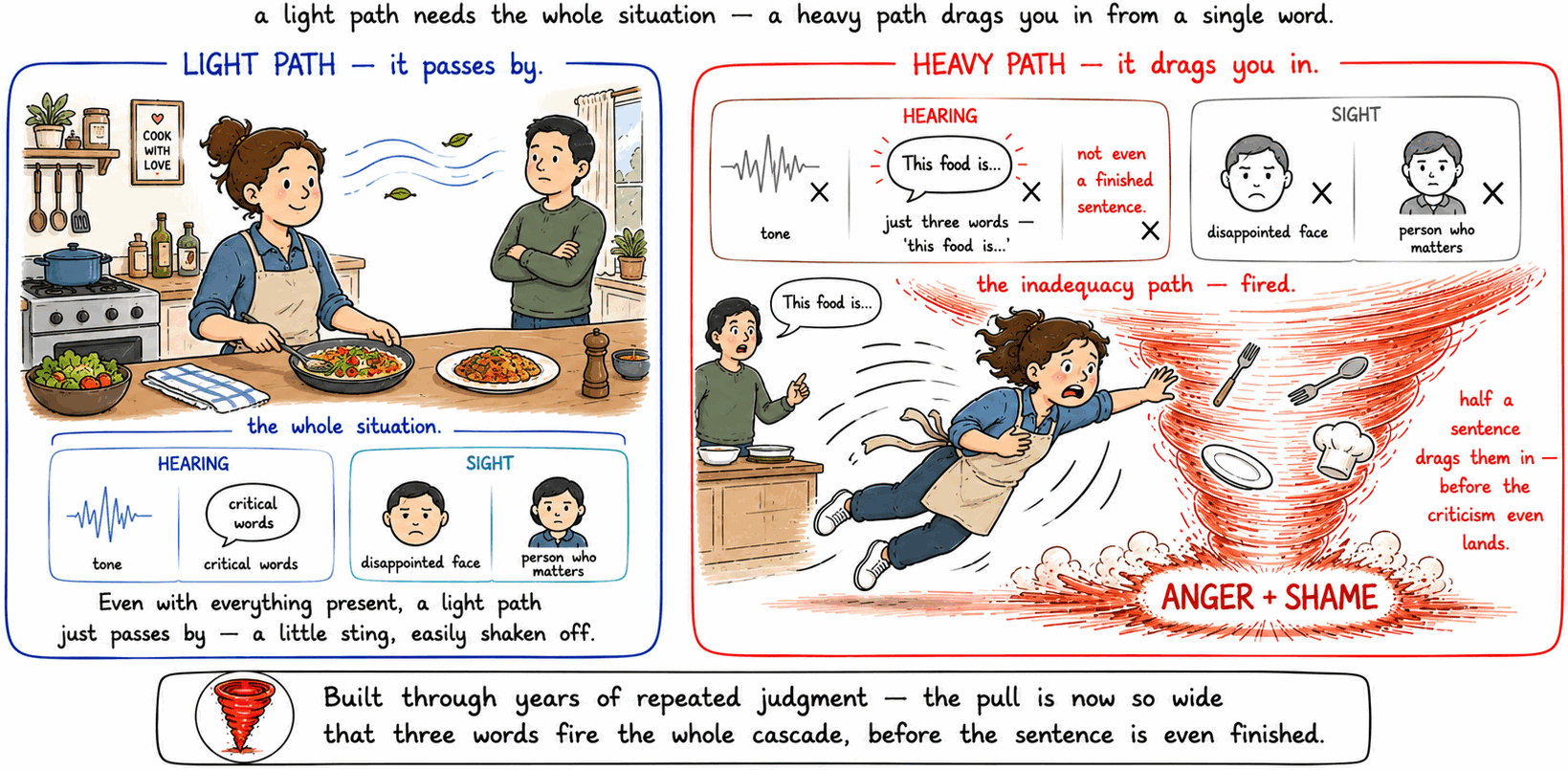}
\caption{The tornado effect in a repetition-encoded path (cook scenario). A 
normally encoded path requires all situational attributes --- tone, facial 
expression, the specific person, and critical words --- to be present to fire 
(left). After deep encoding through years of repeated criticism, the path 
fires from a single attribute: the words ``this food is\ldots'' alone activate 
the full reactive cascade, regardless of tone, speaker, or setting (right).}
\label{fig:tornadocook}
\end{figure}

We propose that the situational specificity required to activate a conditioned 
path is inversely proportional to its encoding depth. A thinly encoded path 
requires the full complement of original situational attributes to fire: the 
tone, the face, the words, and the location must all be present. As encoding 
depth increases, progressively fewer attributes are required, until a heavily 
encoded path fires from a single matching attribute. We term this 
\textit{encoding-depth-dependent stimulus generalisation}, and offer the 
metaphor of a tornado to make it intuitive: a light breeze is felt only on 
direct contact, whereas a tornado exerts its pull far beyond its visible 
boundary~\cite{proof-of-tbi}. The deeper the encoding, the wider the path's 
pull radius (Figure~\ref{fig:tornadocook}).

\begin{figure}[H]
\centering
\includegraphics[width=0.98\textwidth]{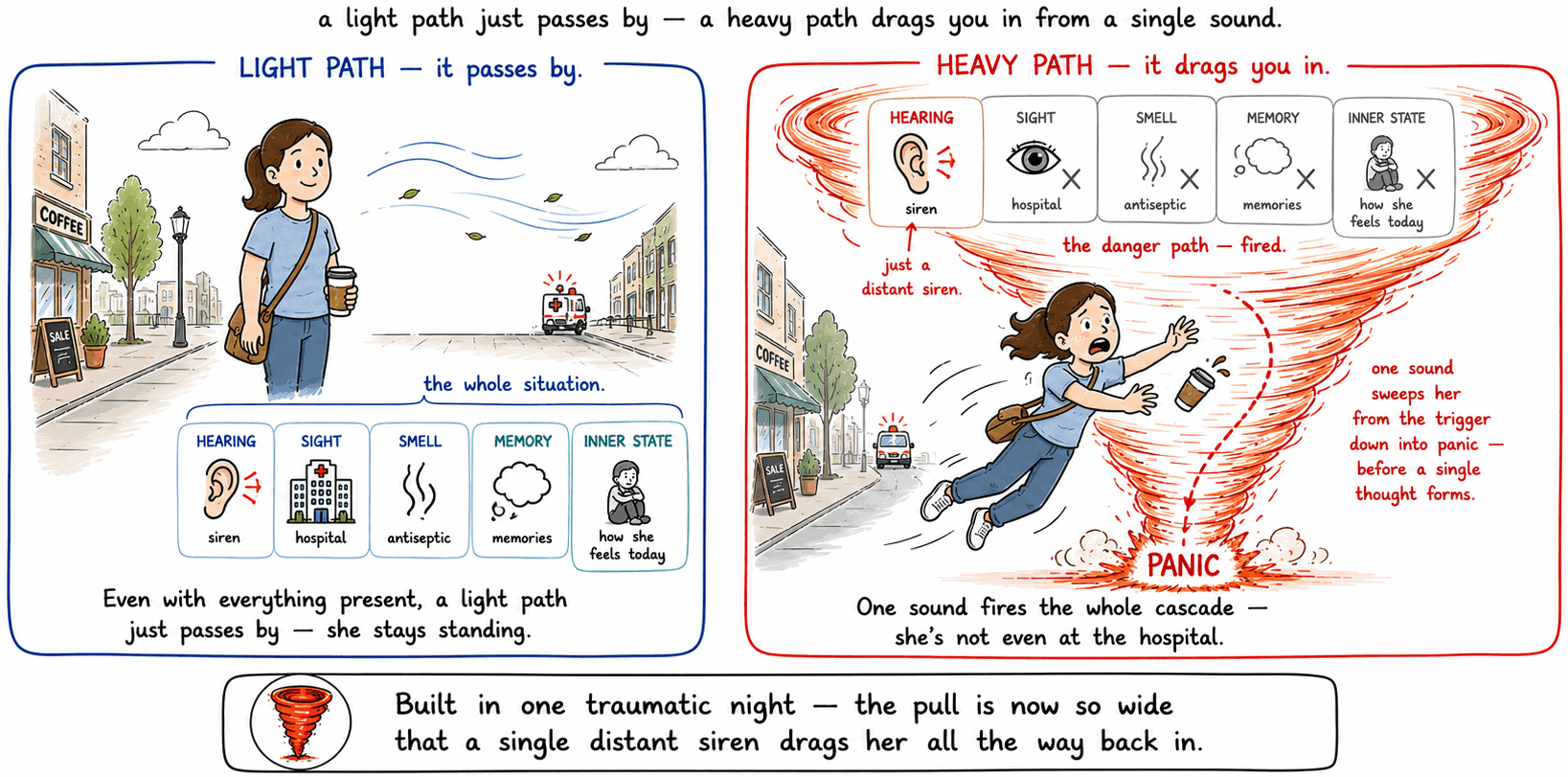}
\caption{The tornado effect in an intensity-encoded path (hospital scenario). 
Following a single mass-casualty event encoded at maximal intensity, the 
danger path acquires an extremely wide pull radius. A distant ambulance siren 
--- a single attribute, absent the original people, place, and circumstances 
--- is sufficient to activate the full reactive cascade. This is the 
mechanistic basis of trauma responses triggering from apparently unrelated 
stimuli.}
\label{fig:tornadoemma}
\end{figure}

The same principle explains the hallmark feature of trauma: activation from 
stimuli that share almost nothing with the original event. When a path is 
encoded at maximal intensity in a single event, its pull radius becomes 
extremely wide, and a single peripheral attribute --- a sound, a smell --- 
suffices to fire the entire cascade \cite{pitman2012, mcnally2003} 
(Figure~\ref{fig:tornadoemma}).

Encoding-depth-dependent stimulus generalisation reframes a puzzling 
observation as a direct consequence of encoding depth, and predicts that 
reducing encoding depth --- narrowing the path --- should correspondingly 
narrow the pull radius, restoring the requirement for more complete 
situational matching before activation.

\subsection{Affective Routing and the Feeling Tone Gap}
\label{subsec:routing}

\begin{figure}[H]
\centering
\includegraphics[width=0.98\textwidth]{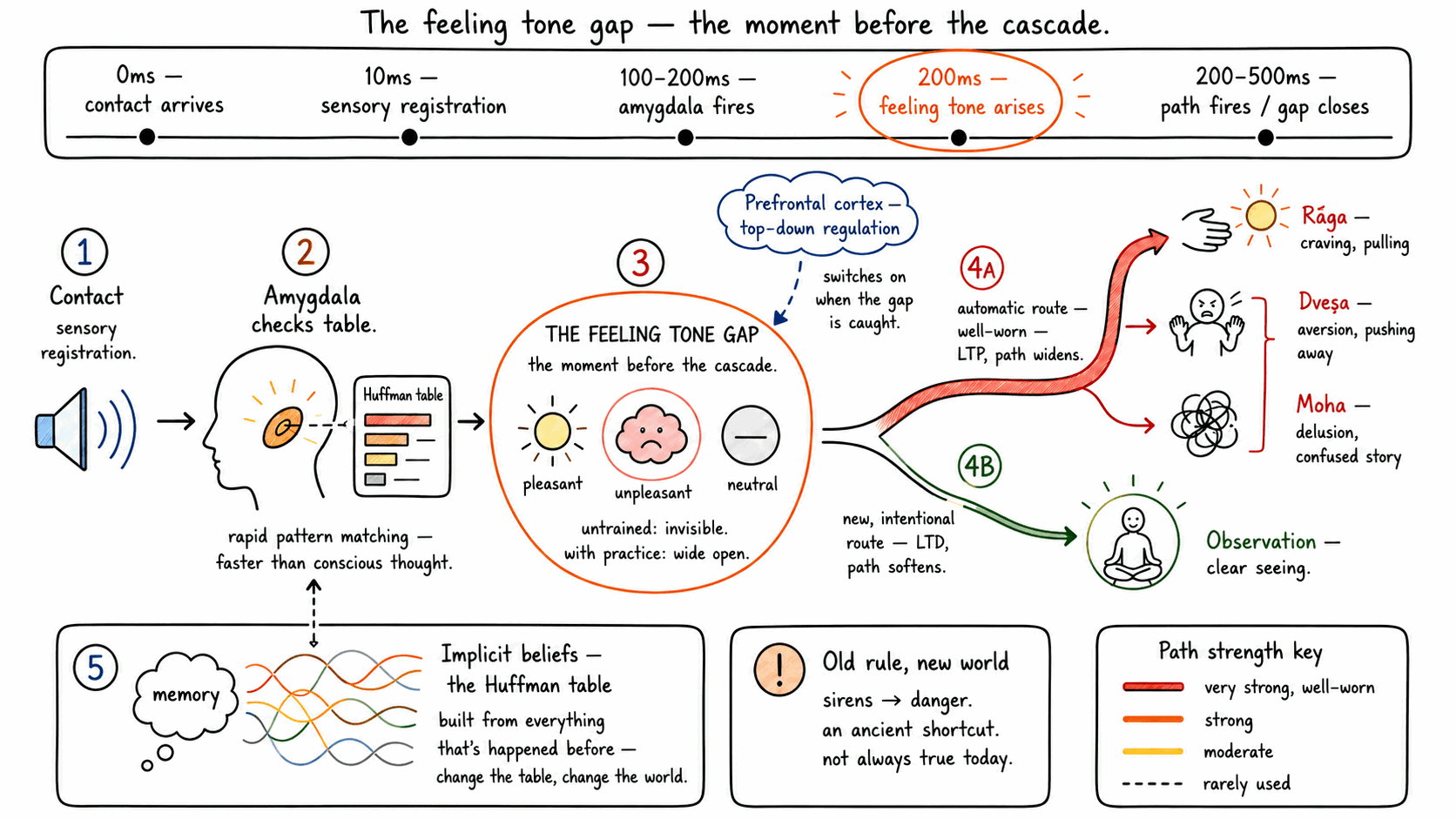}
\caption{How the brain routes thoughts, and the feeling tone gap. A situation 
is registered and matched against encoded paths; the amygdala returns a 
pre-cognitive feeling tone within 100--200ms; and only afterward is the 
conscious thought and reactive behaviour constructed. The brief 
prefrontal-amygdala regulatory window between the feeling tone and the 
completion of the reactive cascade --- the feeling tone gap --- is the sole 
point at which the cascade can be intercepted upstream.}
\label{fig:brainrouting}
\end{figure}

We now specify the temporal structure of activation, because it is here that 
intervention becomes possible. Following situational registration, the 
amygdala generates a pre-cognitive affective signal --- a bare feeling tone of 
pleasantness, unpleasantness, or neutrality --- within approximately 
100--200ms, prior to full prefrontal engagement \cite{ledoux1996, ohman2005, 
zajonc1980}. This feeling tone is not yet an emotion, a thought, or a 
reaction; it is the bare affective valence that precedes and shapes them. The 
constructed emotional experience --- the narrative of what is happening and 
what it means --- is assembled afterward \cite{barrett2017} 
(Figure~\ref{fig:brainrouting}).

Between the arising of the feeling tone and the completion of the reactive 
cascade lies a brief prefrontal-amygdala regulatory window during which 
top-down modulation can intercept the cascade before it completes 
\cite{ochsner2005, gross1998, gross2015, aldao2010}. We call this window the 
\textit{feeling tone gap}. In untrained awareness it is effectively invisible: 
the feeling tone and the reaction appear as a single seamless event. The 
central proposal is that this gap is the only point at which upstream 
intervention --- and therefore structural dissolution --- is possible.

\subsection{How Paths Deepen: Long-Term Potentiation}
\label{subsec:ltp}

\begin{figure}[H]
\centering
\includegraphics[width=0.98\textwidth]{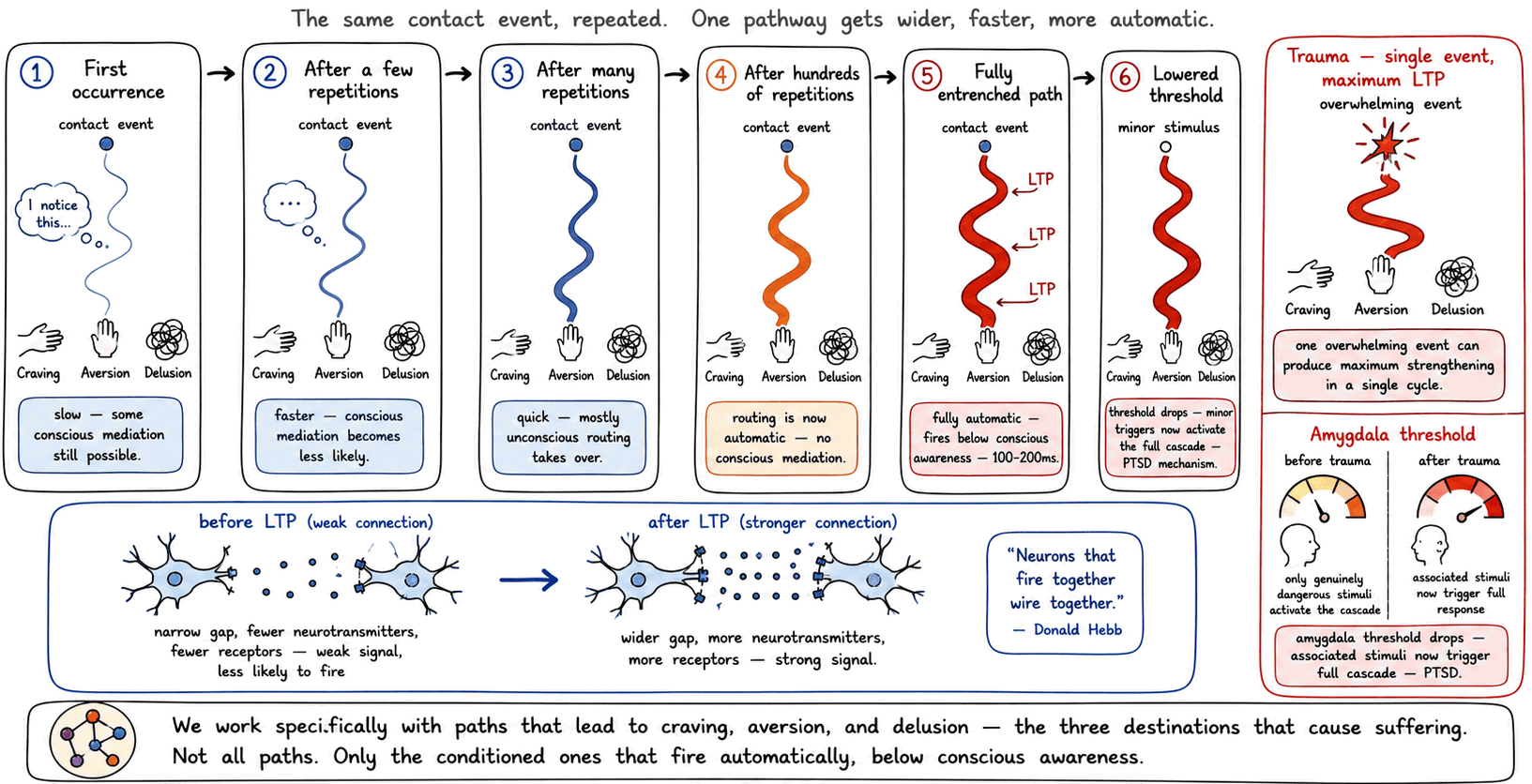}
\caption{How paths deepen through long-term potentiation. Each completed 
firing of a reactive path strengthens its synaptic connections, widening the 
path, lowering its activation threshold, and increasing its firing speed and 
automaticity. Untreated reactive patterns deepen over time because every 
triggered episode functions as a strengthening event.}
\label{fig:pathdeepening}
\end{figure}

Each time a reactive path fires to completion, long-term potentiation operates 
on the synaptic connections along its length: AMPA receptors are inserted into 
the post-synaptic membrane, and the path is strengthened \cite{bliss1993, 
malenka2004, citri2008}. The path becomes wider, its activation threshold 
lower, its firing faster and more automatic, and its pull radius --- following 
the tornado principle --- wider. This is why reactive patterns worsen with 
repetition rather than exhausting themselves, and why avoidance, far from 
relieving a fear, reliably intensifies it: each avoidance episode fires and 
therefore strengthens the very path it seeks to escape \cite{foa2007, 
craske2008} (Figure~\ref{fig:pathdeepening}).

\begin{figure}[H]
\centering
\includegraphics[width=0.98\textwidth]{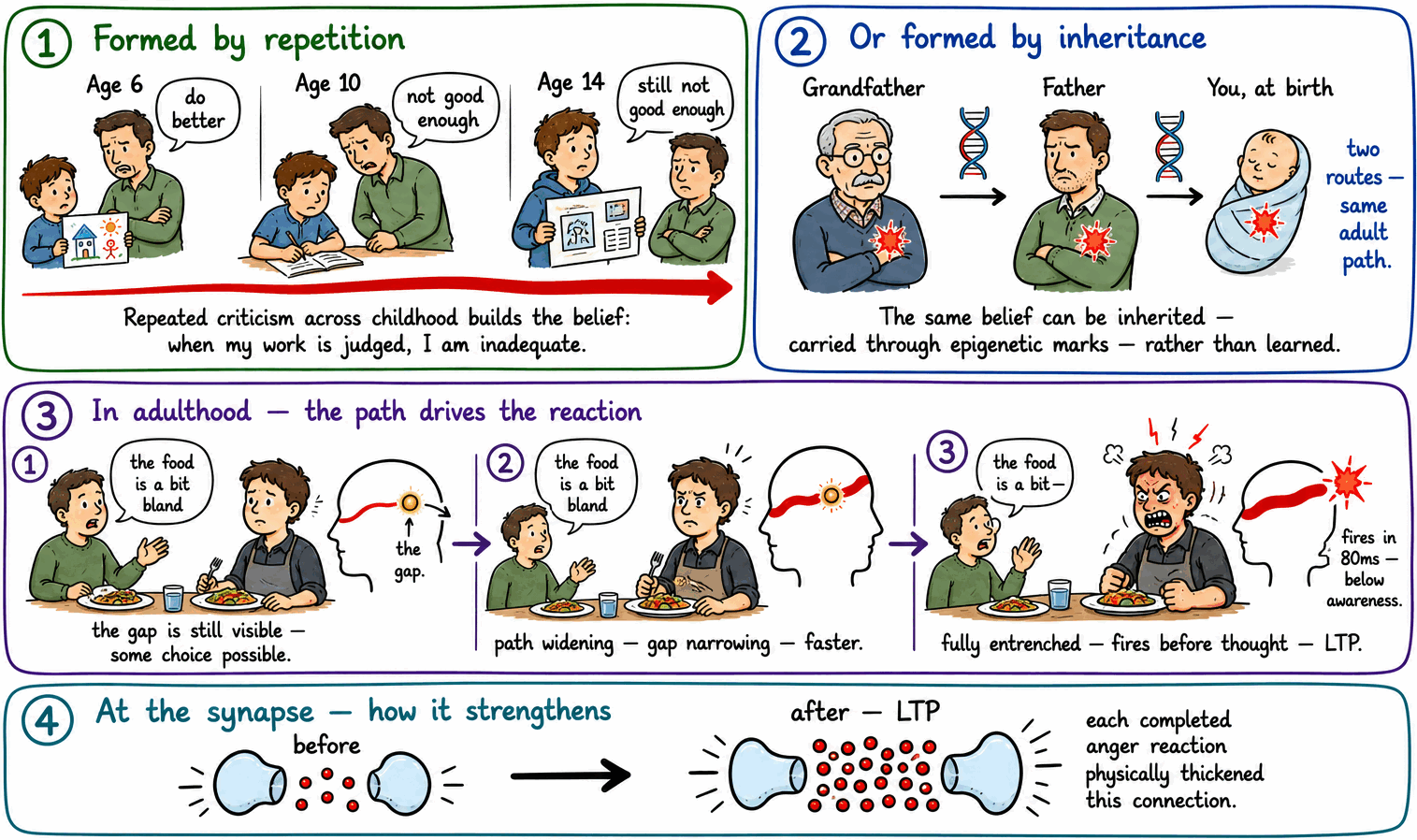}
\caption{Path deepening through repetition (the cook). The inadequacy path is 
built incrementally: each episode of criticism adds a small increment of 
long-term potentiation, and over years these accumulate into a deeply encoded 
path. Every completed firing thereafter deepens it further, widening the path 
and lowering its activation threshold.}
\label{fig:deepeningcook}
\end{figure}

We illustrate path deepening in two contrasting cases, presented separately 
because they represent the two principal formation mechanisms. The first is 
the cook, whose inadequacy path was built by \textit{repetition}. No single 
episode of criticism was decisive; instead, each instance added a 
small increment of long-term potentiation, and over years these increments 
accumulated into a deeply encoded path. Every subsequent completed firing --- 
each adult episode in which criticism of the cook's food fires the path to a 
full reaction --- continues this deepening, widening the path and lowering its 
threshold further (Figure~\ref{fig:deepeningcook}).

\begin{figure}[H]
\centering
\includegraphics[width=0.98\textwidth]{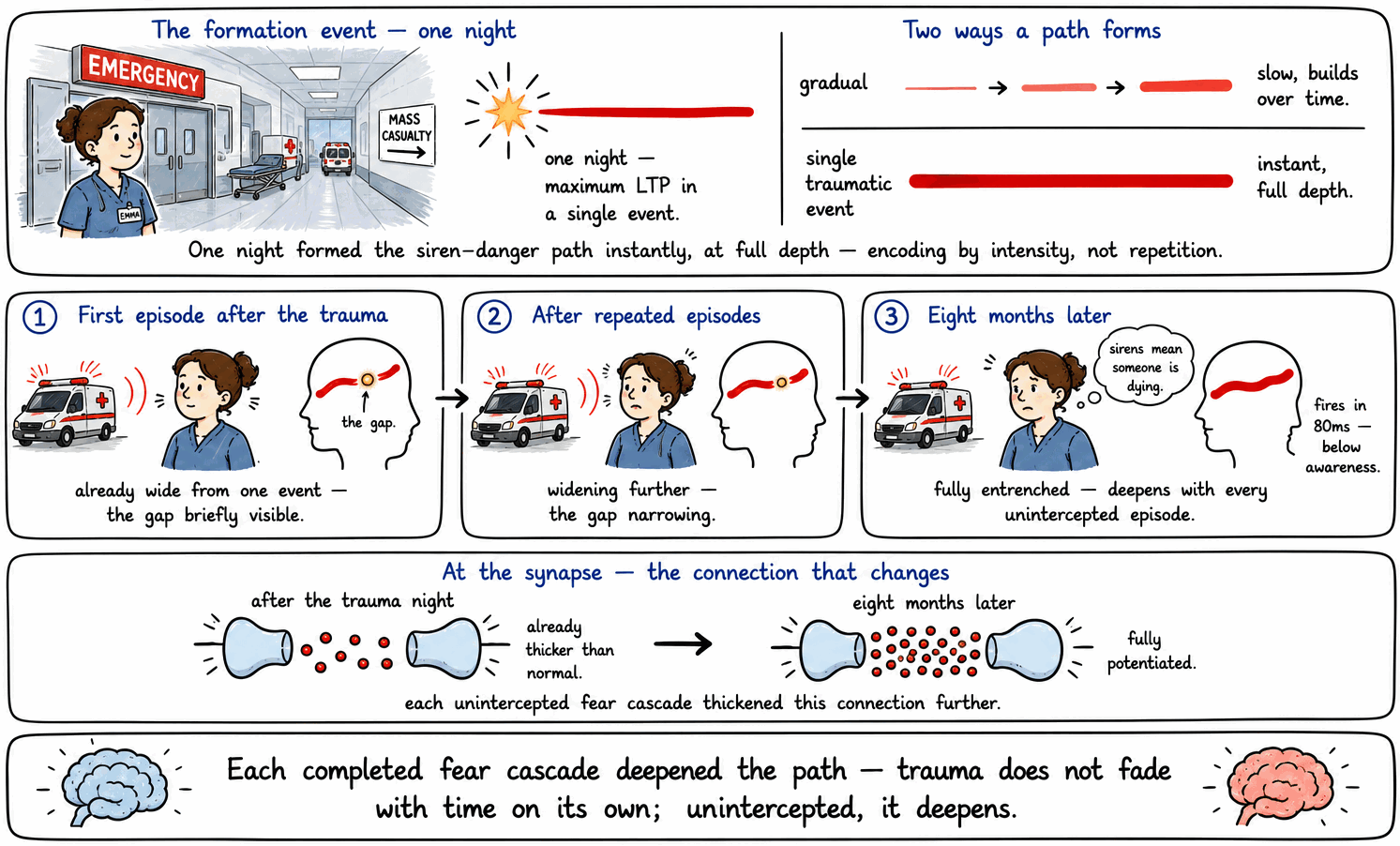}
\caption{Path deepening through single-event intensity (the hospital case, 
Emma). A single night of extreme arousal produces, in one exposure, the 
encoding depth that repetition would require years to build. Each subsequent 
triggered episode of the panic cascade acts as a further potentiation event, 
deepening the path rather than exhausting it.}
\label{fig:deepeninghospital}
\end{figure}

The second is the hospital case (Emma), whose danger path was built by 
\textit{single-event intensity}. Here one overwhelming night produced, in a 
single exposure, the depth of encoding that repetition would take years to 
achieve --- because extreme physiological arousal accelerates synaptic 
consolidation \cite{pitman2012, mcnally2003}. Thereafter every triggered 
episode of the panic cascade operates as a further potentiation event, so 
that untreated the path deepens over time rather than fading 
(Figure~\ref{fig:deepeninghospital}).

The two cases arrive at the same endpoint --- a deeply encoded reactive path 
--- by different temporal routes: many small increments in the cook, one large 
increment in Emma. The implication is identical for both: any intervention 
which allows the path to continue firing to completion --- however it manages 
the aftermath --- leaves the deepening mechanism intact. Structural change 
requires interrupting the firing itself.

\subsection{Which Paths Require Rewiring}
\label{subsec:whichpaths}

\begin{figure}[H]
\centering
\includegraphics[width=0.98\textwidth]{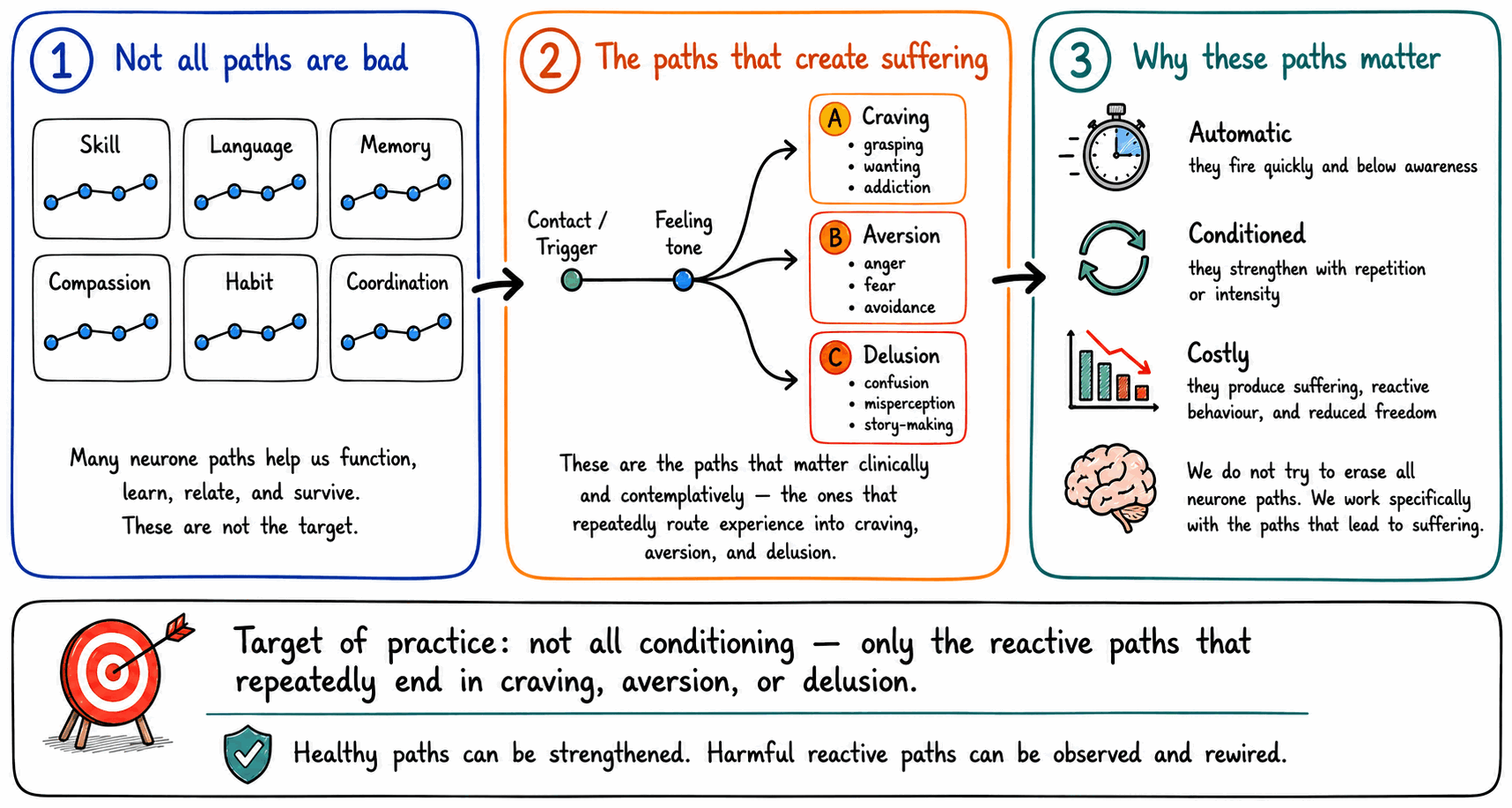}
\caption{Which neurone paths require rewiring. Intervention targets are 
restricted to paths that route automatically to reactive suffering: 
disproportionate, involuntary, and unwanted responses. Adaptive and neutral 
paths are preserved. Identifying the specific target path --- its situational 
key and reactive destination --- is the first step of any intervention built 
on this model.}
\label{fig:whichpaths}
\end{figure}

Because most encoded paths are adaptive, a principled criterion is needed for 
identifying intervention targets. A path is a candidate for rewiring when its 
activation is disproportionate to the situation, involuntary, and productive 
of suffering the person wishes to be free of. Adaptive paths --- skills, 
knowledge, meaningful emotional responses --- are explicitly preserved 
(Figure~\ref{fig:whichpaths}).

Identifying the specific target path, including its situational key and its 
characteristic reactive destination, is the first practical task for any 
intervention grounded in this model.

\subsection{How Paths Dissolve: Long-Term Depression Through Behind-the-Scenes 
Observation}
\label{subsec:ltd}

Long-term depression is the synaptic mechanism opposite to potentiation: under 
conditions of activation without completion, AMPA receptors are internalised 
from the post-synaptic membrane, phosphatases activate, and the synaptic 
connection physically weakens \cite{malenka2004, citri2008, luscher2012, 
bhaskaran2013}. The central mechanistic claim is that engaging the feeling 
tone gap --- observing the path as it begins to fire, rather than allowing it 
to complete --- creates precisely these conditions. Each interception produces 
a small increment of LTD. Individually these increments are negligible; 
accumulated across hundreds of sessions, they produce measurable structural 
reorganisation, narrowing the path, raising its threshold, and reducing its 
pull radius.

We term the practice that achieves this interception \textit{behind-the-scenes 
observation}, and we are precise about what it is and is not. It is not 
calming the feeling, reframing it, replacing it with a more positive thought, 
or accepting it. It is, specifically, the act of consciously watching the 
neural machinery operate in real time as the reaction begins: registering that 
the amygdala has matched the situation and fired, that a conditioned path is 
beginning to route the signal toward its characteristic reactive destination, 
and that if the cascade completes the path will deepen through potentiation~\cite{agentic-workflow-practicle-guide}. 
The content of the observation is mechanistic rather than emotional --- in the 
first person, ``a path is firing here; it was built through past encoding; if 
it completes, the encoding deepens; I am watching this happen; I am not the 
path but its observer.'' We propose that it is this mechanistic quality that 
distinguishes the practice from general awareness and explains its synaptic 
effect: watching one's own reactive architecture operate constitutes an 
unusually deep form of metacognitive engagement, which recruits the strongest 
top-down prefrontal--amygdala regulatory signal and thereby the most complete 
interruption of the cascade before completion \cite{ochsner2005, gross1998, 
gross2015}. In this precise sense, awareness of the mechanism is itself the 
mechanism: it is the seeing of the path, rather than any effort applied after 
the feeling, that creates the activation-without-completion under which LTD 
operates.

We decompose the practice into three layers of observation that trace the 
mechanism's own stages in the order they occur --- signal, feeling tone, and 
neural routing. These layers serve both as a description of what is observed 
and, for people in whom the reactive cascade fires too rapidly to be observed 
directly, as a graded learning scaffold approached one stage at a time. In the 
first layer the person notices the signal arriving --- registering that 
something has been received through one of the six sense bases (sight, 
hearing, smell, taste, touch, or the internal sense of memory), including the 
case in which the signal is an internally arising recollection rather than an 
external event. In the second layer they notice the bare feeling tone the 
signal evokes --- the pre-cognitive valence of pleasantness, unpleasantness, 
or neutrality that arises before any emotion or narrative (``there is 
unpleasantness here'')~\cite{deep-psychiatric}. In the third layer they see the behind-the-scenes 
neural process itself in motion: that the amygdala has matched the signal and 
found a path, that a signal is now routing along that conditioned path toward 
its reactive destination --- craving, aversion, or delusion --- and that if 
the route completes, the path widens and the encoding deepens (``the amygdala 
has found a path; it is routing this toward aversion; if it completes, the 
encoding deepens; I am watching this happen; I am not the path but its 
observer''). The third layer is behind-the-scenes observation proper, and it 
is here that the interception occurs; the first two layers bring the observer 
to it. With practice the three collapse into a single compressed act --- 
registering the signal and its feeling tone and, in the same moment, seeing 
the neural process routing beneath them.

\begin{figure}[H]
\centering
\includegraphics[width=0.98\textwidth]{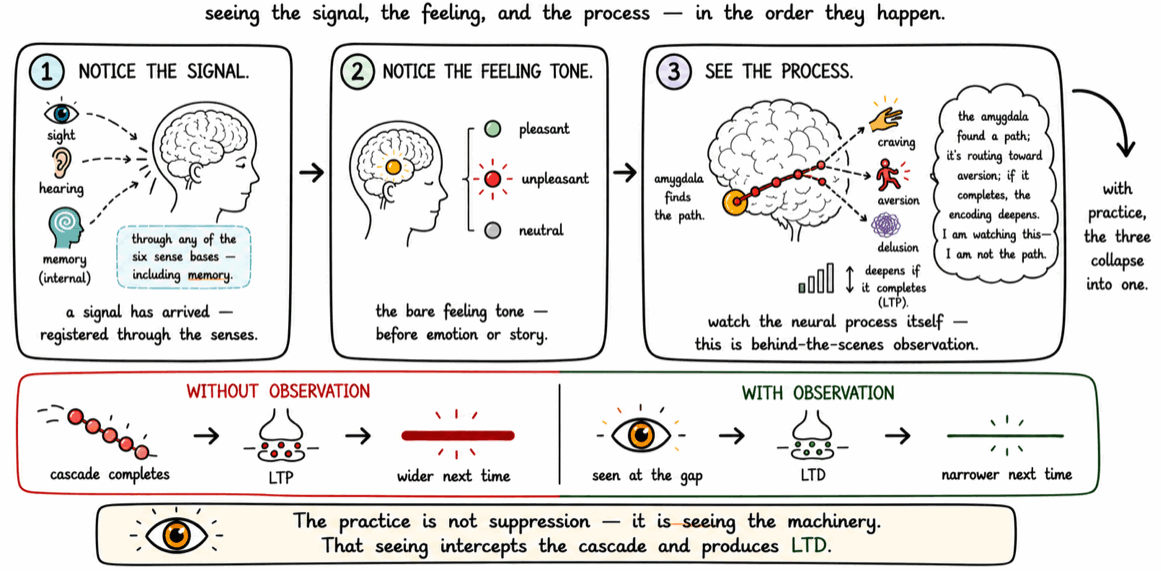}
\caption{How paths dissolve through long-term depression, via the three layers 
of behind-the-scenes observation. The layers trace the mechanism's stages in 
the order they occur: noticing the signal arriving through the senses (Layer 
1), noticing the bare feeling tone it evokes (Layer 2), and seeing the neural 
process itself in motion --- the amygdala matching the signal and a path 
routing toward craving, aversion, or delusion, which deepens if the route 
completes (Layer 3). Seeing the process at Layer 3 intercepts the cascade 
before completion; each interception produces a small increment of long-term 
depression, weakening the path.}
\label{fig:ltdobservation}
\end{figure}

Whereas the three layers describe how deeply the process is seen, a single 
completed interception also has a temporal anatomy: the bare feeling tone is 
noticed as it arises, it is recognised as the activation of a path rather than 
a fact about the world, observation is sustained without allowing the cascade 
to complete, and the path is witnessed subsiding un-acted-upon 
\cite{teasdale1999, fresco2007, wells2009, segal2002}. This gives the 
observation, so often described in general terms, a specific synaptic 
interpretation (Figure~\ref{fig:ltdobservation}).

This mechanism defines the distinction between suppression and dissolution. 
Suppression manages the behavioural output of a cascade that has already fired 
to completion, leaving the path --- and its potentiation --- intact. 
Dissolution intercepts the cascade upstream, at the feeling tone gap, so that 
the path does not fire to completion and long-term depression operates 
instead. Only the latter produces structural change in the path itself.

\begin{figure}[H]
\centering
\includegraphics[width=0.98\textwidth]{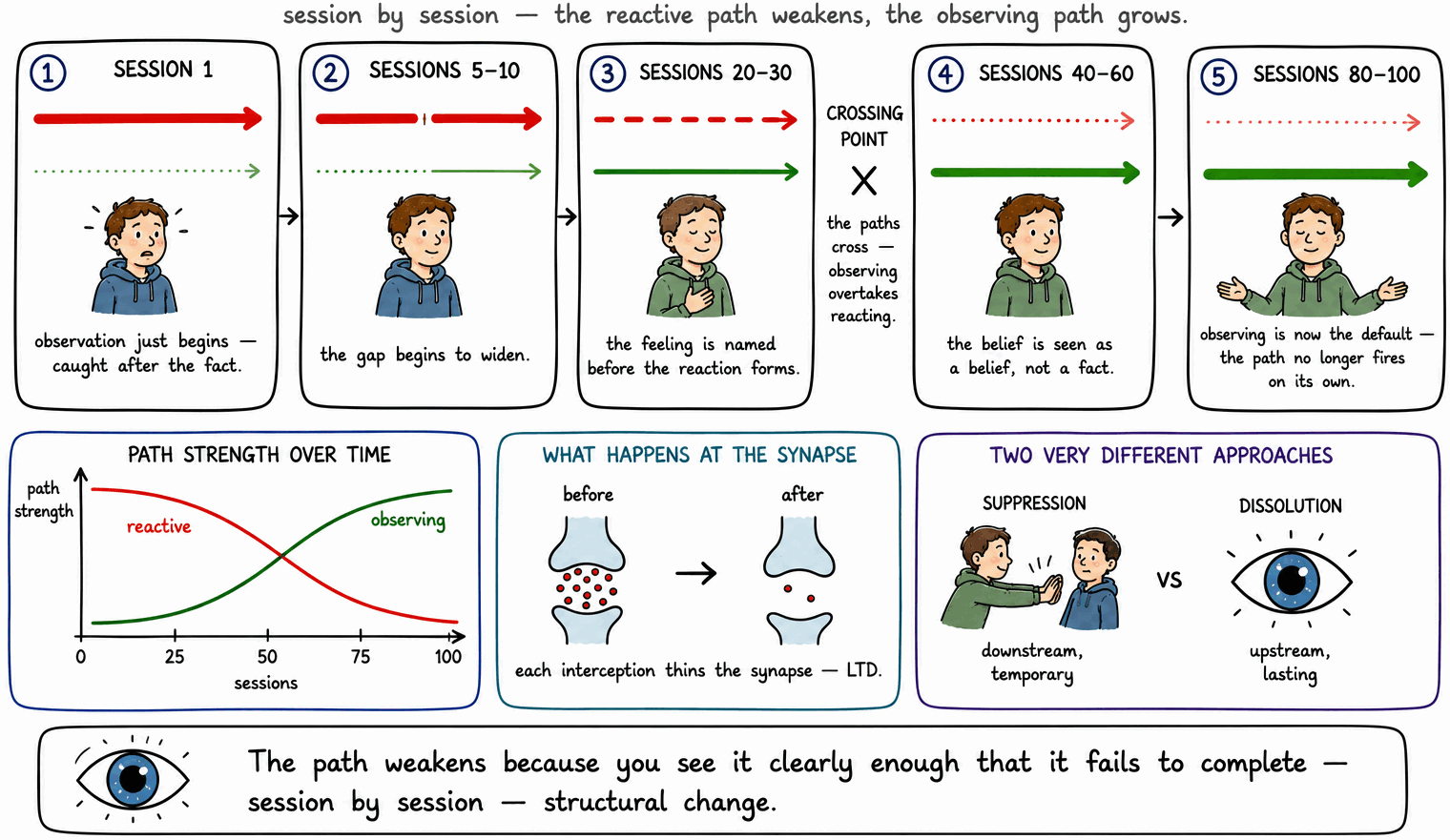}
\caption{How observation rewires the brain across sessions. Repeated 
interception at the feeling tone gap drives two parallel trajectories: the 
reactive path weakens through accumulated long-term depression (its width and 
pull radius contracting session by session), while the observing capacity 
strengthens through potentiation each time it is exercised. Over many sessions 
the trajectories cross --- the observing capacity overtakes the reactive path 
--- after which the situation that once triggered the automatic cascade no 
longer does so. The structural change is cumulative and gradual, not a single 
event.}
\label{fig:observationrewires}
\end{figure}

A single interception changes little; the significance of the mechanism is 
cumulative. Across repeated sessions, two processes run in parallel. The 
reactive path, no longer permitted to fire to completion, undergoes 
incremental long-term depression and progressively weakens. At the same time, 
the act of observation is itself a neural activity, and each time it is 
performed it is strengthened through the same Hebbian potentiation that builds 
any path --- so the capacity to observe grows stronger with practice. Over 
many sessions these opposing trajectories cross: the observing capacity, 
initially weak, becomes stronger than the reactive path, which has narrowed. 
That this kind of repeated mental training produces measurable structural 
change in the brain is well established --- experience-dependent grey-matter 
change follows sustained practice in domains from spatial navigation to 
juggling \cite{maguire2000navigation, draganski2004changes}, and contemplative 
training in particular is associated with structural and functional 
reorganisation in regions supporting attention and emotion regulation 
\cite{holzel2011mindfulness, davidson2003alterations}. Guided observation at 
the feeling tone gap engages precisely this capacity for structural 
reorganisation, directed specifically at the reactive path 
(Figure~\ref{fig:observationrewires}).

\begin{figure}[H]
\centering
\includegraphics[width=0.98\textwidth]{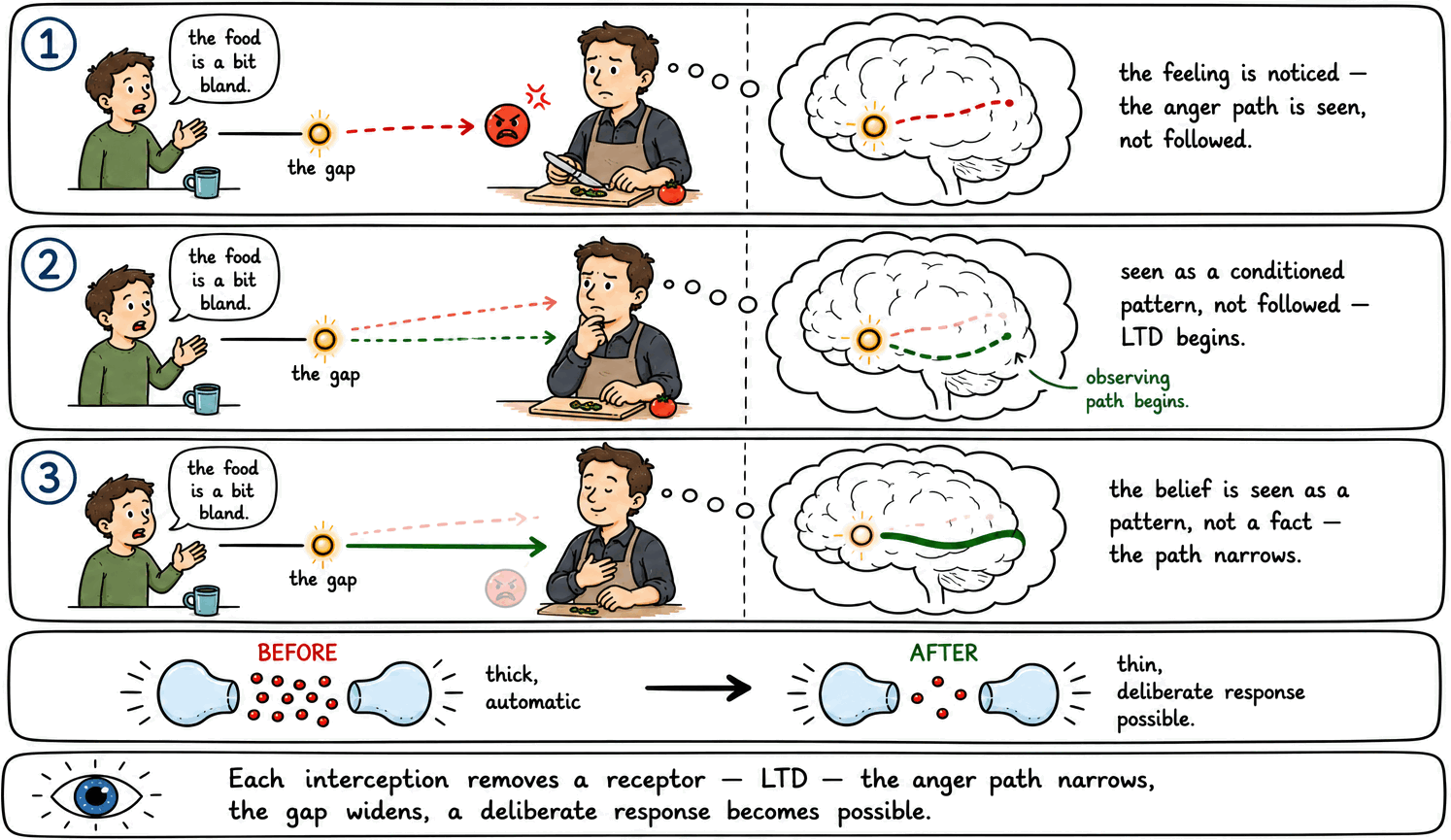}
\caption{Path dissolution through observation (the cook). Repeated observation 
at the feeling tone gap --- catching the ``my work is being judged'' path as 
it begins to fire, rather than completing the reaction --- accumulates 
long-term depression across sessions. The path narrows, its threshold rises, 
and its pull radius contracts, until food criticism no longer fires the full 
reactive cascade automatically.}
\label{fig:ltdcook}
\end{figure}

We illustrate this dissolution process in the same two cases used above. For 
the cook, repeated observation of the ``my work is being judged'' path at the 
feeling tone gap --- catching it as it begins to fire rather than completing 
the reaction --- accumulates long-term depression across sessions, until food 
criticism no longer fires the full cascade automatically 
(Figure~\ref{fig:ltdcook}).

\begin{figure}[H]
\centering
\includegraphics[width=0.98\textwidth]{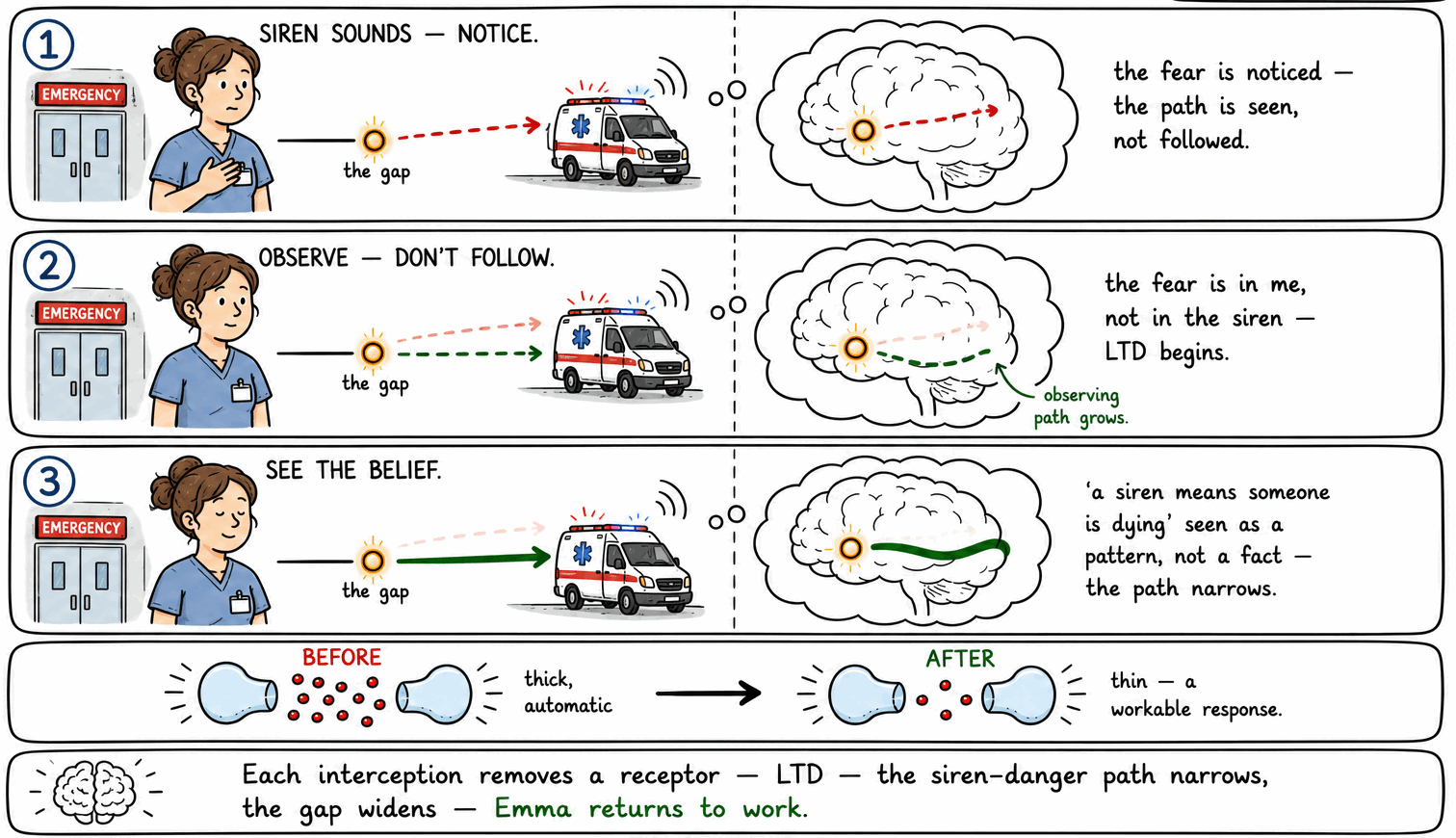}
\caption{Path dissolution through observation (the hospital case, Emma). Though 
the danger path was encoded in a single intense event, it dissolves through 
the same mechanism as the cook's: repeated interception at the feeling tone 
gap accumulates long-term depression, narrowing the path and contracting its 
pull radius, until a siren no longer fires the full panic cascade. The 
encoding history differs; the dissolution mechanism is identical.}
\label{fig:ltdhospital}
\end{figure}

Because the mechanism is defined by interception at the gap rather than by how 
the path was originally encoded, it applies identically to the 
intensity-encoded case. For Emma, repeated observation of the danger path at 
the feeling tone gap --- noticing the panic as it begins to arise, before the 
cascade completes --- accumulates the same long-term depression, session by 
session, narrowing a path that was built in a single night 
(Figure~\ref{fig:ltdhospital}).

Taken together, the components of this section constitute an end-to-end model 
of conditioned reactive patterns: reactive patterns are physical paths; paths 
are encoded by repetition, intensity, and inheritance; a situation reaches the 
brain through a closed set of sensory channels and, as a situation-keyed 
implicit belief, activates a path whose pull radius widens with encoding 
depth; the path fires through a pre-cognitive feeling tone that opens a brief 
regulatory gap; potentiation deepens it when it fires to completion; and 
long-term depression dissolves it when it is intercepted at the gap through 
structured observation. Having established this mechanism with everyday 
examples, we now turn to the setting that is the subject of this paper: how 
these same paths are engaged, and how they may be rewired, during human--AI 
agent interaction (Section~\ref{sec:agents}).

\section{Reactive Paths in Human--AI Agent Environments}
\label{sec:agents}

The mechanism developed in Section~\ref{sec:theory} was illustrated with 
everyday situations --- a criticised cook, a nurse and a siren. This section 
argues that the same mechanism operates, largely unnoticed, in a setting that 
now occupies a growing share of daily life: interaction with AI agents~\cite{astride}. We first characterise the structure of that interaction as an iterative loop, 
then show that each cycle of the loop is a contact event in the precise sense 
of Section~\ref{subsec:situation}, identify which conditioned paths it 
activates, and finally locate within the loop both the potentiation that 
deepens those paths and the feeling tone gap at which they can be intercepted.

\subsection{The Structure of Agent Interaction: The Iterative Loop}
\label{subsec:loop}

\begin{figure}[H]
\centering
\includegraphics[width=0.98\textwidth]{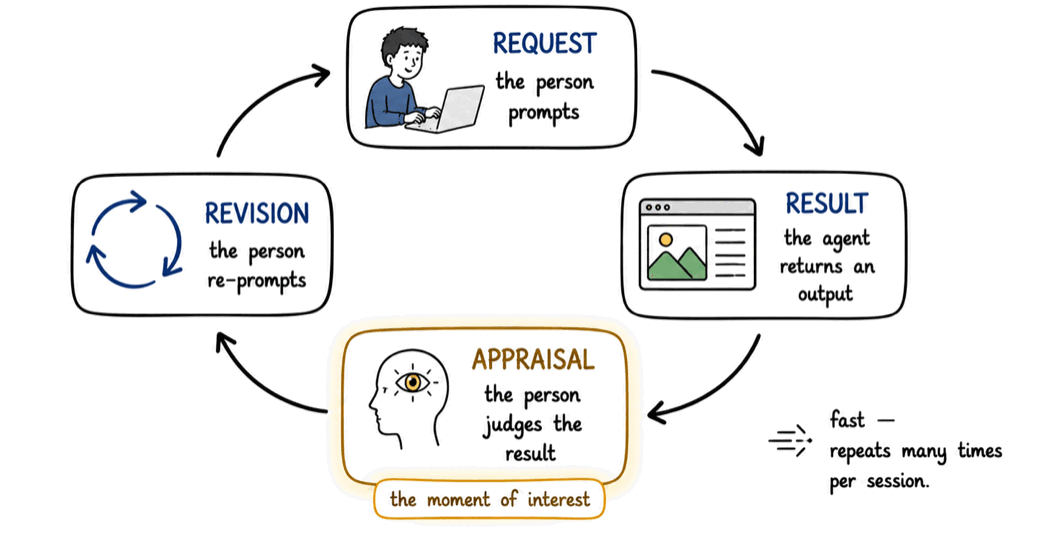}
\caption{The iterative structure of human--AI agent interaction. The person 
issues a request; the agent returns a result; the person appraises the result; 
and the person revises and re-issues the request. This four-phase loop --- 
request, result, appraisal, revision --- repeats rapidly, many times within a 
single task. The appraisal phase, in which the returned result meets the 
person, is the point of interest for this paper.}
\label{fig:interactionloop}
\end{figure}

Interaction with an AI agent --- a conversational assistant, a coding 
copilot, or a generative tool for images, text, or design --- proceeds through 
a characteristic iterative loop. The person issues a request; the agent 
returns a result; the person evaluates that result; and, on the basis of that 
evaluation, revises the request and issues it again. A prompt is written, an 
output appears, the output is judged, and a new prompt follows~\cite{agentsway, agentic-ai}. This 
four-phase cycle --- request, result, appraisal, revision --- repeats many 
times within a single task and many times again across a working session 
(Figure~\ref{fig:interactionloop}).

Two properties of this loop distinguish it from most everyday situations in 
which reactive patterns are encountered, and both will matter for the argument. 
First, it is \textit{high-frequency}: where the situations of 
Section~\ref{sec:theory} might recur a few times a day, the interaction loop 
can complete dozens of times within a single session, each completion 
delivering a fresh result to be appraised. Second, the interval between 
stimulus and response is \textit{short and sharply bounded}: the result 
appears at a definite moment, and the reactive response --- the revised 
prompt, often typed in evident frustration --- follows within seconds. The 
loop thus compresses into minutes a volume of stimulus--response cycles that, 
in ordinary life, would be spread across days.

\subsection{Each Cycle Is a Contact Event}
\label{subsec:contactevent}

\begin{figure}[H]
\centering
\includegraphics[width=0.98\textwidth]{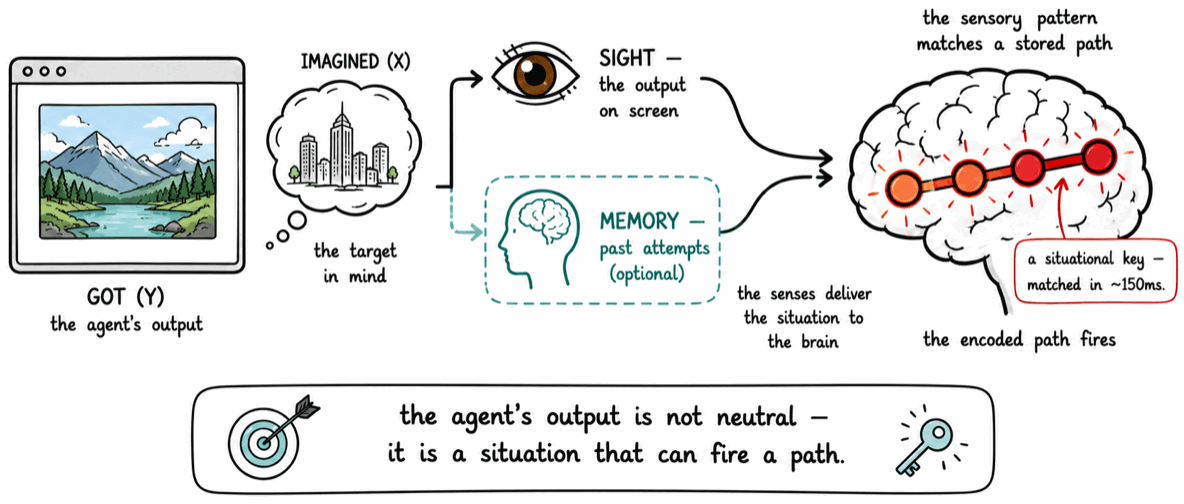}
\caption{An agent's result as a contact event. When the agent returns a 
result --- here, a generated image that does not match what was wanted --- the 
result reaches the person through their senses (predominantly sight) and, 
optionally, through the internal channel of memory (the recalled history of 
previous attempts). This pattern of sensory attributes is a situational key, 
matched by the amygdala against encoded paths exactly as any other situation 
is. The agent's output is thus not a neutral stimulus but a situation capable 
of firing a conditioned reactive path.}
\label{fig:agentcontactevent}
\end{figure}

The appraisal phase of the loop --- the moment at which the returned result 
meets the person --- is a contact event in the exact sense established in 
Section~\ref{subsec:situation}: a situation reaches the person through their 
sensory channels and is matched against encoded paths. When a generated image 
appears on screen, the situation arrives predominantly through sight; when a 
coding agent's output breaks a running program, it may arrive through sight 
together with an internal channel --- the recalled history of similar failures. 
The returned result is, in the terms of Section~\ref{subsec:situationpath}, a 
situational key: a pattern of sensory attributes that the amygdala matches, 
within approximately 100--200ms, against the person's store of encoded paths 
\cite{ledoux1996, ohman2005} (Figure~\ref{fig:agentcontactevent}).

This identification is the pivot of the paper. It means that the agent's 
output is not a neutral technical artefact but a situation in the full sense 
of Section~\ref{sec:theory} --- and therefore that it can, and routinely does, 
activate conditioned reactive paths. Because the loop is high-frequency, it 
supplies these contact events in dense succession. Ordinary agent use is thus 
a continuous stream of situations, each capable of firing a path, arriving far 
faster than in most other domains of life.

\subsection{Which Paths the Agent Fires}
\label{subsec:whichfire}

The paths that agent interaction activates are not new or agent-specific. They 
are the same conditioned paths a person already carries, now keyed by a novel 
class of situation. Two are especially common, and both were foreshadowed by 
the everyday examples of Section~\ref{sec:theory}.

The first is the path that treats \textit{one's own output as being judged}. 
When a person prompts an agent and receives a result that falls short, the 
situation --- one's effort, evaluated, found wanting --- is structurally the 
same as the one that fires the cook's inadequacy path in 
Section~\ref{subsec:situationpath}. To a brain that carries a deeply encoded 
``my work is being judged'' path, a disappointing generated image is not 
merely a technical outcome; it is an instance of the very situation the path 
was built to detect. The evaluation the person performs on the agent's output 
doubles, below awareness, as an evaluation felt to be performed on the person~\cite{think-before-you-act, responsible-ai}.

The second is the path that concludes \textit{one cannot do this}. When 
repeated prompts fail to produce the intended result, the situation --- 
sustained effort at a task without success --- can fire an encoded inadequacy 
or helplessness path of exactly the kind laid down, in Section~\ref{subsec:formation}, 
by a childhood of being told one's efforts are not good enough. The agent 
becomes the occasion for a much older reactive pattern.

\begin{figure}[H]
\centering
\includegraphics[width=0.98\textwidth]{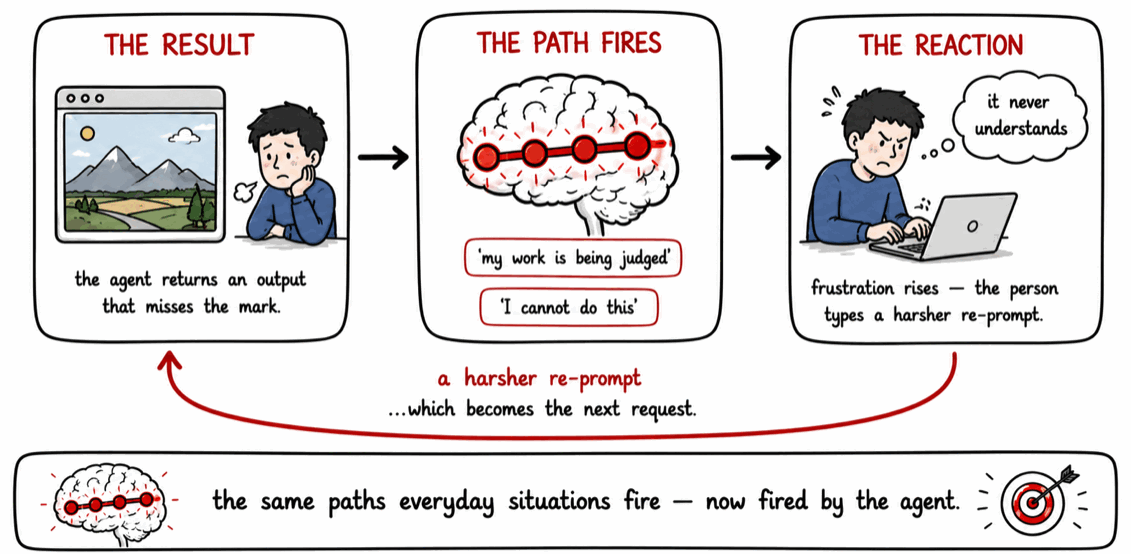}
\caption{An agent's result firing a reactive path. A disappointing result 
(left) reaches the person as a situational key and matches a deeply encoded 
path --- here, ``my work is being judged'' or ``I cannot do this'' (centre). 
The path fires to completion: a pre-cognitive feeling tone of unpleasantness, 
a constructed narrative, and a reactive behaviour --- a harsher, more 
frustrated re-prompt (right). The same paths that everyday situations fire 
(Section~\ref{sec:theory}) are fired here by the agent's output.}
\label{fig:agentpathfires}
\end{figure}

Alongside these, agent interaction reliably evokes reactive paths of 
impatience (the demand that the result arrive now), of perfectionism (the 
refusal to accept any result short of an unspecified ideal), and of control 
(the felt need for the tool to comply exactly)~\cite{explainable-ai-text}. In each case the reactive 
response completes in the familiar way: a feeling tone of unpleasantness, a 
constructed narrative (``it never understands,'' ``I am wasting my time,'' ``I 
can't even do this''), and a reactive behaviour --- an abrupt, often harsher 
re-prompt (Figure~\ref{fig:agentpathfires}).

The reactive behaviour is worth dwelling on, because it closes the loop. The 
frustrated re-prompt is not outside the interaction; it is the next request, 
which produces the next result, which is appraised in turn. The completion of 
the reactive cascade is thus fed directly back into the loop as its next 
input.

\subsection{The Loop as a Potentiation Engine}
\label{subsec:looppotentiation}

\begin{figure}[H]
\centering
\includegraphics[width=0.98\textwidth]{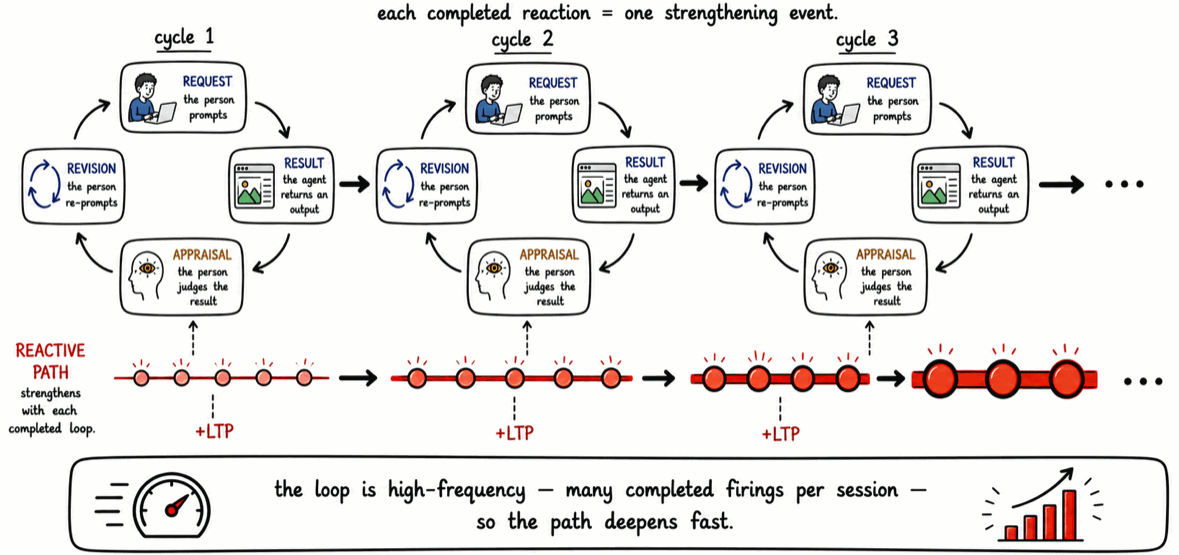}
\caption{The interaction loop as a potentiation engine. Each cycle in which 
the agent's result fires a reactive path to completion deposits an increment 
of long-term potentiation, widening the path and lowering its threshold. 
Because the loop is high-frequency --- completing many times within a single 
session --- these increments accumulate rapidly. Repeated, unobserved agent 
interaction can therefore deepen a reactive path faster than most everyday 
situations do.}
\label{fig:agentltp}
\end{figure}

Section~\ref{subsec:ltp} established that each time a reactive path fires to 
completion, long-term potentiation strengthens it: the path widens, its 
threshold falls, and it fires faster and more automatically the next time 
\cite{bliss1993, malenka2004, citri2008}. The interaction loop, left 
unobserved, is therefore a potentiation engine. Each cycle in which a 
disappointing result fires the ``my work is being judged'' path to completion 
is one more strengthening event; and because the loop is high-frequency, these 
events accumulate far faster than in ordinary life. A single frustrating 
prompting session may complete the reactive cascade a dozen times or more, 
each completion depositing a small increment of potentiation on the same path 
(Figure~\ref{fig:agentltp}).

This yields an observation with some weight. Hundreds of millions of people 
now spend substantial time each day inside this loop \cite{brynjolfsson2023generativeai, 
noy2023experimental}. To the extent that the loop repeatedly fires reactive 
paths of impatience, perfectionism, and self-judgment to completion, everyday 
agent interaction may be quietly training precisely those 
patterns --- not because the technology is designed to, but because the 
structure of the loop, combined with the ordinary human tendency to react to 
disappointing results, makes each session a dense sequence of potentiation 
events. The same property that makes the loop a training ground for reactive 
patterns, however, is what makes it available for the opposite purpose.

\subsection{The Feeling Tone Gap Within the Loop}
\label{subsec:loopgap}

\begin{figure}[H]
\centering
\includegraphics[width=0.98\textwidth]{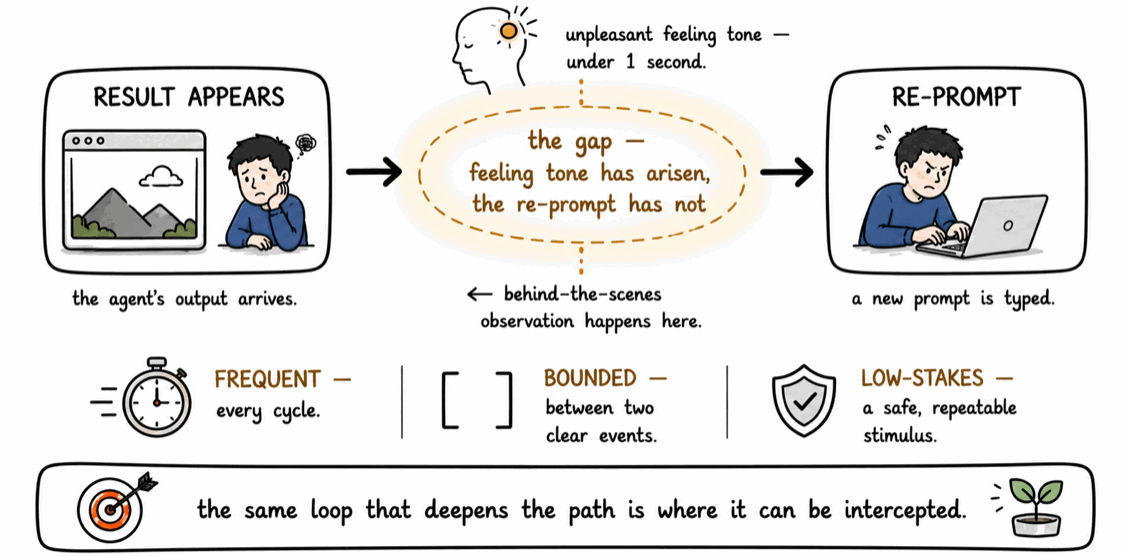}
\caption{The feeling tone gap within the interaction loop. The gap falls at 
the appraisal phase, in the brief interval between the result appearing and 
the reactive re-prompt being issued. Because both the stimulus (the result) 
and the response (the re-prompt) are discrete, observable events, the gap is 
comparatively easy to locate here --- more so than in the seamless situations 
of ordinary life. This bounded, repeatable, low-stakes gap is what makes the 
interaction loop suitable as a site of intervention, developed in 
Section~\ref{sec:rewiring}.}
\label{fig:loopgap}
\end{figure}

Section~\ref{subsec:routing} located the sole point of upstream intervention 
at the feeling tone gap: the brief window between the arising of the 
pre-cognitive feeling tone and the completion of the reactive cascade. Within 
the interaction loop, this gap has an unusually definite location. It falls at 
the appraisal phase --- in the moment between the result appearing and the 
reactive re-prompt being issued~\cite{aitrust-os}. When a generated image resolves on screen and 
is not what was wanted, there is a brief interval, typically under a second, in 
which a feeling tone of unpleasantness has arisen but the frustrated re-prompt 
has not yet been typed. That interval is the feeling tone gap, and in the 
interaction loop it is comparatively easy to locate, because the stimulus (the 
result appearing) and the response (the re-prompt) are both discrete, 
observable, external events with a clear temporal boundary between them 
(Figure~\ref{fig:loopgap}). It is at this interval that the behind-the-scenes 
observation described in Section~\ref{subsec:ltd} would be performed --- 
watching the amygdala match the disappointing result to the ``my work is being 
judged'' path, and the path begin to route toward frustration, rather than 
completing that route by firing off the reactive re-prompt.

Three features make this gap unusually favourable as a site of practice. It is 
\textit{frequent}, recurring with every cycle of a high-frequency loop, so 
opportunities to engage it are abundant rather than rare. It is 
\textit{bounded}, falling between two discrete external events, so it is easier 
to locate than the gap in a fluid interpersonal situation. And it is 
\textit{low-stakes}: a disappointing image or an imperfect draft is a safe, 
repeatable, inconsequential stimulus, which makes the interaction loop a 
gentler training context than the high-stakes situations --- criticism from a 
superior, a traumatic reminder --- in which reactive paths are usually 
encountered. The loop that Section~\ref{subsec:looppotentiation} identified as 
a potentiation engine is, for exactly the same structural reasons, an unusually 
tractable site for the interception that produces long-term depression. How 
that interception is performed --- by the person alone, or with the agent's 
assistance --- is the subject of Section~\ref{sec:rewiring}.

\section{Rewiring Through Agent Interaction}
\label{sec:rewiring}

Section~\ref{sec:agents} established two facts about the interaction loop that 
now converge. Left unobserved, the loop is a potentiation engine: it fires 
reactive paths to completion at high frequency and thereby deepens them 
(Section~\ref{subsec:looppotentiation}). Yet the same loop contains, at its 
appraisal phase, an unusually well-defined feeling tone gap 
(Section~\ref{subsec:loopgap}). The proposal of this section is direct: if the 
behind-the-scenes observation defined in Section~\ref{subsec:ltd} is performed 
at that gap, in place of the reactive re-prompt, the very same loop that 
deepened the path begins instead to dissolve it. The loop does not need to be 
abandoned or replaced; it needs only to be engaged differently at one point in 
its cycle.

\subsection{Turning the Loop from Potentiation to Depression}
\label{subsec:turningloop}

Consider the ordinary case. A generated image resolves on screen and is not 
what was wanted; a feeling tone of unpleasantness arises within a fraction of 
a second; and the reactive cascade completes as a frustrated re-prompt, which 
fires the path and deposits its increment of potentiation 
(Section~\ref{subsec:whichfire}). The reactive re-prompt is the completion of 
the cascade.

\begin{figure}[H]
\centering
\includegraphics[width=0.98\textwidth]{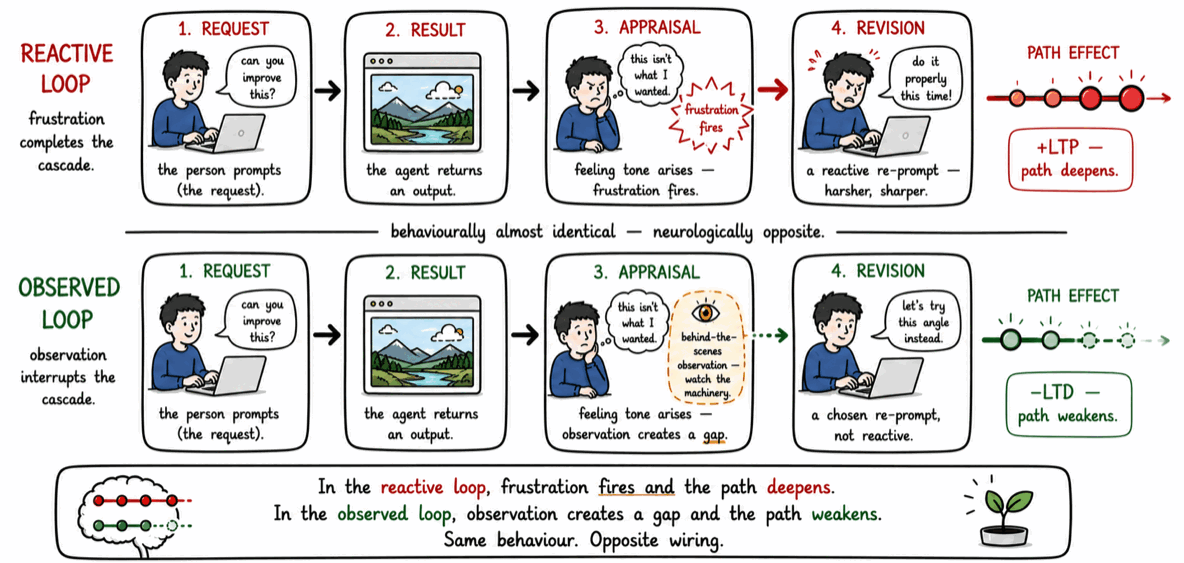}
\caption{Turning the interaction loop from potentiation to depression. In the 
reactive loop (top), a disappointing result fires the path to completion as a 
frustrated re-prompt, depositing long-term potentiation and deepening the 
path. In the observed loop (bottom), behind-the-scenes observation is performed 
at the feeling tone gap instead: the reaction is watched rather than completed, 
the conditions for long-term depression are met, and the path weakens. The 
task still proceeds --- a revised prompt is still issued --- but it is chosen 
from a position of observation rather than issued reactively.}
\label{fig:looprewire}
\end{figure}

The intervention substitutes, at the gap, a different act. Rather than allowing 
the feeling tone to complete into the reactive re-prompt, the person performs 
behind-the-scenes observation: they watch the machinery operate --- the 
disappointing result has been matched to the ``my work is being judged'' path; 
the path is beginning to route toward frustration; were the frustrated 
re-prompt to follow, the path would deepen. Observed in this way, the cascade 
does not complete. The conditions for long-term depression --- activation 
without completion --- are met, and the path receives a small increment of 
weakening rather than strengthening (Section~\ref{subsec:ltd}). Crucially, the 
person still revises the prompt; the task proceeds. But the revised prompt is 
now issued from a position of having observed the reaction rather than from 
within it --- chosen rather than reactive (Figure~\ref{fig:looprewire}).

The two loops are behaviourally almost identical --- in both, a result appears 
and a revised prompt follows --- but they are neurologically opposite. The 
difference is entirely a matter of what happens at the gap: whether the feeling 
tone completes unobserved into a reactive re-prompt, or is observed before the 
re-prompt is issued. This is the whole of the intervention, and its smallness 
is the point: it asks not for a change of tool, task, or workflow, but for a 
single change of stance at one recurring moment.

\subsection{Learning the Practice: The Three-Layer Scaffold on the Loop}
\label{subsec:scaffoldloop}

Behind-the-scenes observation is a skill, and like any skill it is built 
through repetition (Section~\ref{subsec:formation}). Section~\ref{subsec:ltd} 
described the graded three-layer scaffold through which it is learned. The 
interaction loop is an unusually favourable place to climb that scaffold, for 
precisely the three properties identified in Section~\ref{subsec:loopgap}: the 
gap recurs with every cycle, so the learner has abundant repetitions; it is 
bounded between two discrete events, so it is easy to locate; and it is 
low-stakes, so the practice can proceed without the reaction being 
overwhelming.

\begin{figure}[H]
\centering
\includegraphics[width=0.98\textwidth]{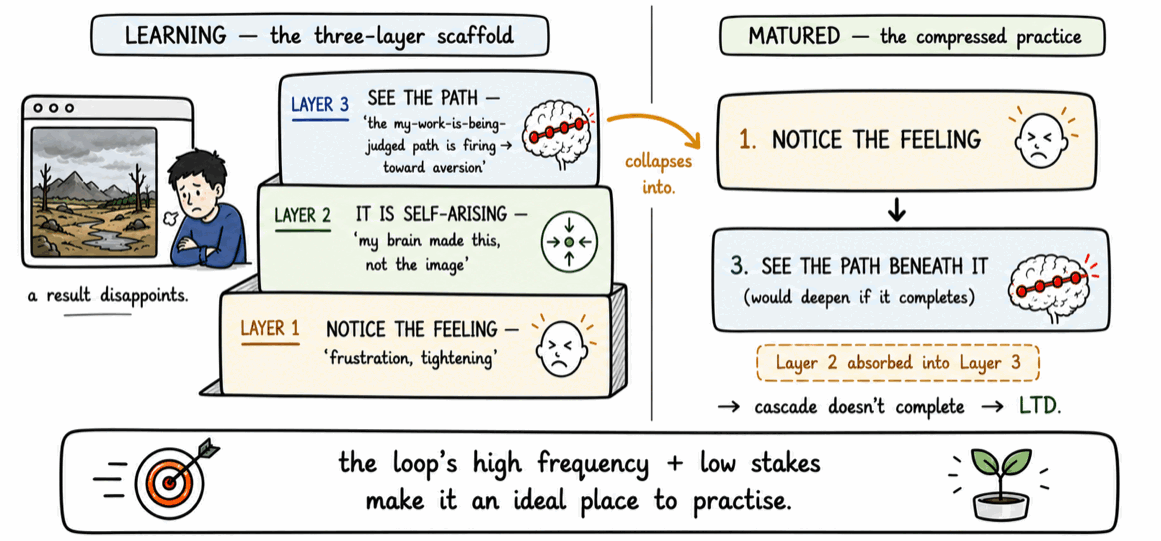}
\caption{Learning behind-the-scenes observation on the interaction loop, via 
the three layers, and its compressed form. The three layers trace the 
mechanism's stages as they occur. During learning (left), when a result 
disappoints: Layer 1 --- notice the signal arriving (the result through sight, 
and past attempts through the internal sense of memory); Layer 2 --- notice 
the bare feeling tone it evokes (``frustration, tightening''); Layer 3 --- see 
the neural process in motion (``the amygdala has found the `my work is being 
judged' path; a signal is routing toward aversion; it deepens if it 
completes''). Once Layer 3 is reliable, the layers collapse into the compressed 
practice (right): register the signal and its feeling tone and, in the same 
moment, see the process routing beneath them. The loop's high frequency and 
low stakes make it an ideal setting to practise.}
\label{fig:scaffoldloop}
\end{figure}

Climbing the scaffold in this setting proceeds through the three layers of 
Section~\ref{subsec:ltd}, which trace the mechanism's stages as they occur. In 
the first layer, when a result disappoints, the learner notices the signal 
arriving --- the disappointing image registering through sight (and, often, a 
recalled history of previous failed attempts arriving through the internal 
sense of memory). In the second layer they notice the bare feeling tone the 
signal evokes --- ``there is frustration here, a tightening'' --- before 
typing anything. In the third layer they see the behind-the-scenes neural 
process in motion: that the amygdala has matched this result to the ``my work 
is being judged'' path and a signal is now routing along it toward aversion, 
and that were the route to complete in a frustrated re-prompt, the path would 
widen and the encoding deepen. The third layer names, in the terms of 
Section~\ref{subsec:huffman}, the reactive destination toward which the path 
routes --- craving, aversion, or delusion --- and it is here, in seeing the 
process rather than completing it, that the interception occurs. With practice 
the three layers collapse, as described in Section~\ref{subsec:ltd}, into the 
single compressed act of behind-the-scenes observation 
(Figure~\ref{fig:scaffoldloop}).

In its mature, compressed form the practice is a brief movement in which the 
signal and its feeling tone are registered and, in the same moment, the neural 
process beneath them is seen. When a result disappoints and the feeling 
arises, the practised observer sees the neurone path routing beneath it --- a 
conditioned path that, if it fires to completion, routes toward aversion and 
deepens its own encoding. That seeing is itself the interception: the cascade 
does not complete, and long-term depression operates. Expanded into its full 
content, the compressed observation runs approximately: 
\textit{``something is happening in my brain right now; the amygdala has fired; 
there is a conditioned path here, built through past encoding, routing this 
toward aversion; if I complete that route the path widens and the encoding 
deepens; I am watching this happen.''} In the compressed form the separate 
registration of signal and feeling tone is no longer a deliberate step; the 
whole movement resolves into the single act of seeing the neural process 
routing beneath the reaction.

The low-stakes character of the setting, which might appear to make it trivial, 
is in fact its central advantage as a training ground. A disappointing 
generated image is an inconsequential stimulus: nothing depends on it, it can 
be reproduced at will, and the reactive path it fires is a mild instance of the 
same path that fires in far more consequential situations. This makes the loop 
an ideal place to build the observing capacity cheaply and repeatedly on 
low-cost material. And because the reactive path fired by a disappointing 
result is the \textit{same} path --- ``my work is being judged'' --- that fires 
when a superior criticises one's work or a reviewer rejects one's submission, 
the observing capacity built on the trivial case is the same capacity that 
becomes available in the consequential one. The loop is a low-stakes gymnasium 
for a skill that transfers to high-stakes situations.

\subsection{Mode A: User-Guided Practice}
\label{subsec:modea}

In the user-guided mode, a person who understands the framework practises 
behind-the-scenes observation during their own ordinary agent use, with no 
change to the tools they already use. The practice requires no application, no 
configuration, and no additional software; it requires only that the person, 
at the moment a result disappoints them, engage the gap as described rather 
than completing the reactive re-prompt. The agent and its interface are 
unchanged; what changes is the stance the person brings to the appraisal phase 
of a loop they were already performing. This mode has the advantage of being 
available immediately and universally to anyone who interacts with any agent, 
and the limitation of depending entirely on the person's own developed capacity 
to remember, at the moment of reaction, to observe rather than react --- a 
capacity that is initially weak, and is exactly what the scaffold is designed 
to build.

\subsection{Mode B: Agent-Assisted Practice}
\label{subsec:modeb}

The limitation of the user-guided mode motivates the second. Because the 
person is already interacting with an agent at the precise moment the reactive 
path fires, the agent is uniquely positioned to support the observation --- not 
as a bespoke clinical system, but as an ordinary agent lightly configured to do 
so. The configuration required is minimal: at the level of a system prompt or 
an optional interaction mode, an agent can be instructed to recognise the 
signs of a reactive re-prompt --- a terse, frustrated, or self-critical 
revision --- and to respond not merely by complying but by briefly inviting the 
person to observe what has just arisen before continuing 
\cite{bandara2025responsible}. A single unobtrusive prompt at the right moment 
--- \textit{``before we revise: notice what came up when that result 
appeared''} --- can convert a reactive re-prompt into an occasion for 
observation, supplying externally the reminder that the user-guided mode 
depends on the person to generate internally.

\begin{figure}[H]
\centering
\includegraphics[width=0.98\textwidth]{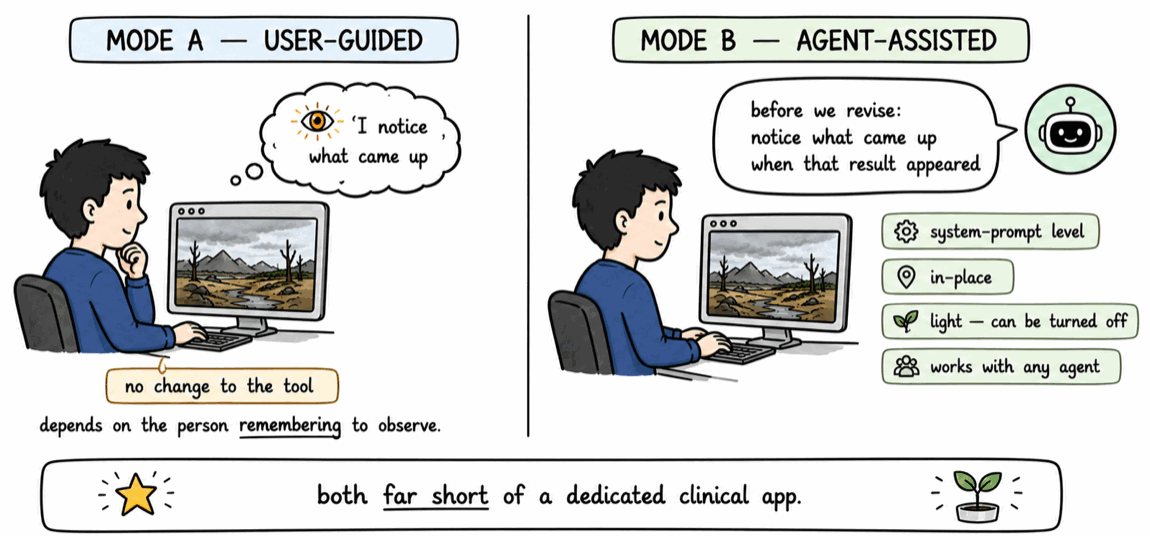}
\caption{Two modes of practice within agent interaction. In the user-guided 
mode (left), the person performs behind-the-scenes observation at the gap 
unaided, during ordinary agent use, with no change to the tool. In the 
agent-assisted mode (right), an ordinary agent is lightly configured --- at 
the level of a system prompt --- to notice a reactive re-prompt and briefly 
invite observation before continuing. The agent-assisted mode supplies 
externally the reminder that the user-guided mode depends on the person to 
generate, while remaining far short of a dedicated clinical application.}
\label{fig:twomodes}
\end{figure}

Several design considerations distinguish this from a dedicated therapeutic 
application. The assistance is \textit{in place}: it occurs within the tool the 
person is already using, at the moment of reaction, rather than in a separate 
session after the fact. It is \textit{light}: it does not diagnose, does not 
store a clinical record, and can be declined or disabled at any time. And it is 
\textit{general}: the same minimal configuration can be applied to any agent 
--- a coding copilot, a writing assistant, a generative image tool --- because 
the reactive structure it responds to is the same across all of them 
(Figure~\ref{fig:twomodes}). The agent-assisted mode thus occupies a deliberate 
middle ground: more supportive than unaided self-practice, but far short of a 
purpose-built clinical system, and available within tools people already use.

\subsection{Why Agent Interaction Is Well-Suited to This Purpose}
\label{subsec:wellsuited}

The properties that recommend agent interaction as a site for this practice can 
now be drawn together. It is \textit{high-frequency}: the loop supplies many 
opportunities to practise within a single session, where ordinary life supplies 
few. It is \textit{immediate and bounded}: the gap falls between two discrete, 
observable events, making it easier to locate than the gap in a fluid 
interpersonal exchange. It is \textit{low-stakes}: the triggering stimulus is 
safe and repeatable, so the observing capacity can be built cheaply. It is 
\textit{co-located}: uniquely, the reactive moment and the opportunity to 
practise occur in the same place, at the same time, within the same tool --- 
so no transfer from a separate practice setting to the moment of reaction is 
required, because they are one and the same. And, through the agent-assisted 
mode, it admits \textit{in-place support} at the exact moment of reaction, 
which no unaided practice and no after-the-fact intervention can offer. These 
properties do not make agent interaction a replacement for other approaches to 
reactive patterns; they make it a newly available, and unusually convenient, 
site at which the same underlying mechanism of behind-the-scenes observation 
can be practised. Section~\ref{sec:example} develops a concrete case --- a 
generative image-prompting session --- in which these abstractions become a 
specific, recognisable sequence of events.

\section{Use Case: Generative Image Prompting}
\label{sec:example}

The preceding sections developed the framework in general terms. This section 
grounds it in a single concrete case: a session of generative image prompting. 
We choose this case not because it is the most consequential setting in which 
reactive paths fire --- it is among the least --- but because it makes the 
mechanism unusually easy to see. We first say why, then follow one short 
session twice: once unobserved, as it ordinarily unfolds, and once with 
behind-the-scenes observation performed at each gap.

\subsection{Why Image Prompting Makes the Mechanism Visible}
\label{subsec:whyimage}

Three properties of generative image prompting make it an ideal illustrative 
case. It is \textit{fast}: a result returns within seconds, so a single 
session compresses many complete cycles of the interaction loop into a few 
minutes, and the potentiation or depression of a path across repetitions 
becomes observable within one sitting. It is \textit{visual}: the mismatch 
between what was intended and what was produced is apprehended immediately and 
wholly, without reading or interpretation, so the feeling tone it evokes 
arrives with particular sharpness. And it is \textit{emotionally salient}: 
because the image is a direct rendering of the person's own expressed 
intention, a result that falls short is readily felt as a small verdict on the 
person --- which is precisely the situation that fires the ``my work is being 
judged'' path (Section~\ref{subsec:whichfire}). These properties make the 
prompting session a natural microcosm in which the abstractions of the 
preceding sections become a specific, recognisable sequence of events.

\subsection{The Session Unobserved}
\label{subsec:unobserved}

Consider a person prompting an image tool to produce a specific picture they 
have in mind. The first result returns, and it is wrong in some obvious way --- 
the composition is not what they asked for. A feeling tone of mild 
unpleasantness arises, below deliberate thought, and completes into a reaction 
before it is noticed: a quick, slightly terser re-prompt, issued from 
impatience --- the demand that the result simply arrive. The path has fired to 
completion; a small increment of potentiation is deposited 
(Section~\ref{subsec:looppotentiation}).

The second result is closer but still not right. Now the feeling tone is 
sharper, and the reaction that completes is one of perfectionism: the result 
is rejected not against the original intention but against an ideal that has 
crept upward, and the re-prompt is issued with an edge. The third result 
introduces a new flaw while fixing the old one, and here a different path 
fires: a thought arrives, fully formed and unbidden --- ``I can't even 
describe what I want; I am not good at this'' --- which is the ``I cannot do 
this'' path of Section~\ref{subsec:whichfire}, an old inadequacy pattern that 
the tool has merely occasioned. The re-prompt that follows is now driven by a 
felt need to force the tool into compliance, the reactive pattern of control.

Across a dozen such cycles, each disappointing result fires a reactive path to 
completion, and each completion deepens it. The session ends with the person 
more frustrated than they began, a worse image than they hoped for, and --- the 
point that matters here --- several reactive paths each a little more deeply 
encoded than before. Nothing about the session was unusual; this is simply how 
such sessions ordinarily go. And that is exactly the concern of 
Section~\ref{subsec:looppotentiation}: performed this way, at this frequency, 
the loop is efficiently training impatience, perfectionism, self-judgment, and 
control.

\subsection{The Same Session Observed}
\label{subsec:observed}

Now consider the same session --- the same prompts, the same disappointing 
results --- conducted by a person practising behind-the-scenes observation at 
each gap.

\begin{figure}[H]
\centering
\includegraphics[width=0.98\textwidth]{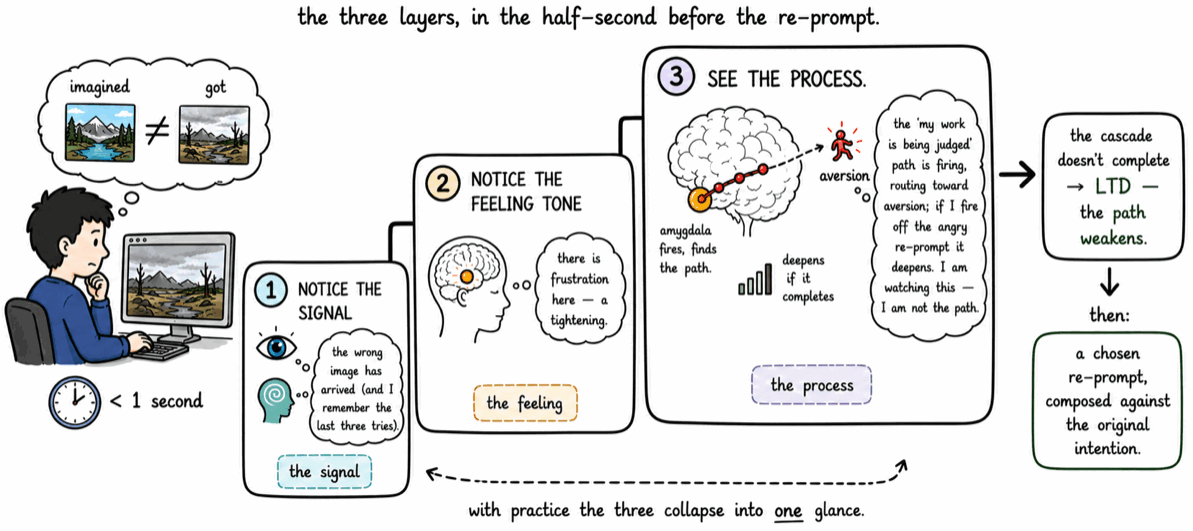}
\caption{The three layers of behind-the-scenes observation applied to a single 
disappointing result, in the half-second before the re-prompt. Layer 1: notice 
the signal that has arrived (the wrong image through sight, and the recalled 
previous attempts through the internal sense of memory). Layer 2: notice the 
bare feeling tone it evokes (a small unpleasantness, a tightening). Layer 3: 
see the neural process in motion --- the ``my work is being judged'' path 
routing toward aversion, which would deepen were the frustrated re-prompt 
issued. Seen in this way the cascade does not complete, long-term depression 
operates, and the revision that follows is chosen rather than reactive. With 
practice the three layers collapse into a single glance.}
\label{fig:observedmoment}
\end{figure}

The first result returns wrong. The feeling tone of unpleasantness arises 
exactly as before; but this time, in the interval before the re-prompt, it is 
observed. Walked through the three layers of Section~\ref{subsec:ltd}, the 
observation unfolds as follows. In the first layer the person notices the 
signal that has arrived: the wrong image on the screen, registered through 
sight, together with the recalled history of the previous failed attempts 
arriving through the internal sense of memory. In the second layer they notice 
the bare feeling tone the signal evokes --- a small unpleasantness, a 
tightening --- before it becomes a story. In the third layer they see the 
neural process itself in motion: \textit{the ``my work is being judged'' path 
has been matched and is beginning to route toward aversion; this result is an 
arrangement of pixels, and the sting is a conditioned response of mine, not a 
property of the image; were I to complete this into a frustrated re-prompt, the 
path would deepen.} With practice these three collapse into a single glance, 
the compressed practice of Section~\ref{subsec:scaffoldloop} 
(Figure~\ref{fig:observedmoment}). Seen in this way, the cascade does not 
complete. The conditions for long-term depression are met, and the path 
weakens by a small increment (Section~\ref{subsec:ltd}). The person then 
revises the prompt --- the task proceeds exactly as before --- but the revision 
is chosen, composed against the original intention rather than issued from 
impatience.

The second and third results, still disappointing, are met the same way. When 
the ``I cannot do this'' thought arrives at the third result, it too is 
observed rather than believed: \textit{this is the old inadequacy path firing, 
occasioned by a difficult prompt; it is a conditioned encoding, not a fact 
about my ability.} Observed, it does not complete into the spiral of 
self-criticism it would otherwise have driven. Across the same dozen cycles, 
each disappointing result becomes an occasion not for potentiation but for a 
small increment of depression, and the reactive paths --- impatience, 
perfectionism, self-judgment, control --- each weaken slightly rather than 
deepen.

\subsection{What the Two Sessions Show}
\label{subsec:whatdiffers}

\begin{figure}[H]
\centering
\includegraphics[width=0.98\textwidth]{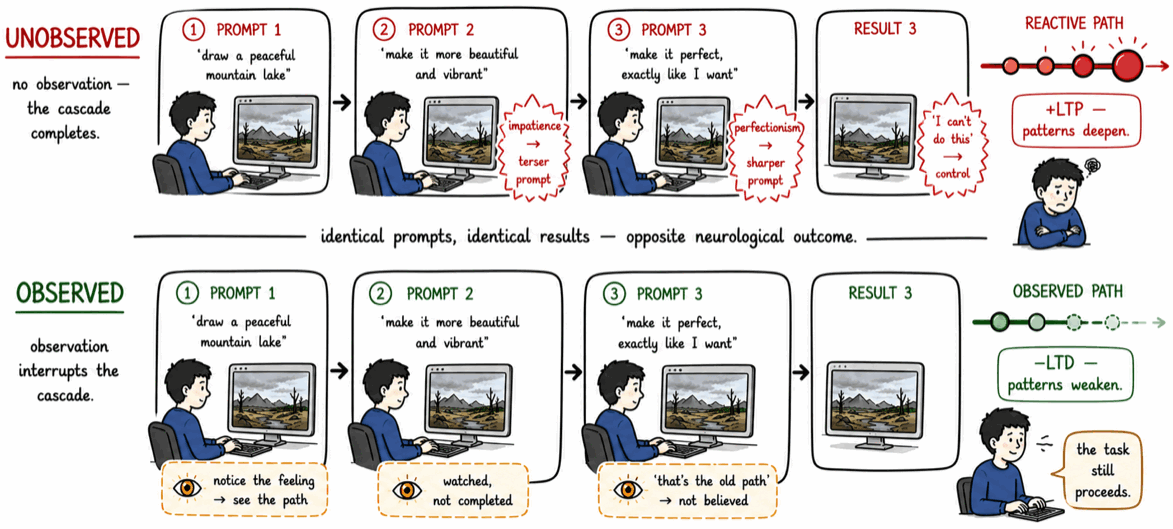}
\caption{The same prompting session, unobserved and observed. Both sessions 
share identical prompts and identical disappointing results. In the unobserved 
session (top), each result fires a reactive path --- impatience, perfectionism, 
self-judgment, control --- to completion, depositing long-term potentiation and 
deepening the pattern. In the observed session (bottom), behind-the-scenes 
observation is performed at each gap: the feeling is seen as a firing path 
rather than completed into a reaction, depositing long-term depression and 
weakening the pattern. The visible behaviour is nearly identical; the 
neurological outcome is opposite.}
\label{fig:promptingsession}
\end{figure}

The two sessions are, on the surface, nearly the same: in both, a person 
prompts an image tool, receives a series of disappointing results, and revises 
repeatedly. An observer watching the screen might not easily tell them apart. 
Yet neurologically they are opposite. In the first, a dozen contact events 
each fired reactive paths to completion, depositing potentiation and deepening 
the patterns. In the second, the same dozen contact events were each 
intercepted at the gap, depositing depression and weakening the patterns. The 
entire difference lay in what happened in the sub-second interval between each 
result and each re-prompt --- whether the feeling tone completed unobserved, or 
was seen (Figure~\ref{fig:promptingsession}).

Two observations follow. First, the prompting session is a microcosm: a dozen 
repetitions of exactly the contact-event cycle described in 
Section~\ref{sec:agents}, compressed into minutes, in which the effect of 
observation versus non-observation on a reactive path can be seen in a single 
sitting. Second, and more consequentially, the paths practised upon in this 
trivial setting are not trivial paths. The ``my work is being judged'' path 
weakened slightly over a frustrating image session is the same path that fires 
when the person's actual work is criticised by someone whose opinion matters; 
the observing capacity built cheaply on disappointing images is the same 
capacity that becomes available in that harder case 
(Section~\ref{subsec:scaffoldloop}). The worked example is therefore not merely 
an illustration of the mechanism but an instance of the training the framework 
proposes: a low-stakes, high-frequency setting in which a transferable capacity 
is built on inconsequential material.

\section{Discussion}
\label{sec:discussion}

The framework developed here makes a specific and, we acknowledge, ambitious 
proposal: that the ordinary loop of human--AI agent interaction is a site at 
which conditioned reactive patterns are both deepened and, engaged 
differently, dissolved. We discuss here what this implies for the design of 
agentic systems, how it relates to established approaches, its limitations, 
and the studies that would test it.

\subsection{Design Implications for Agent Builders}
\label{subsec:design}

The most immediate implication follows from an observation made in 
Section~\ref{subsec:looppotentiation}: agents shape their users' reactive 
patterns whether or not anyone intends them to. Every agent that returns 
results a user reacts to is already participating in the potentiation or 
attenuation of that user's reactive paths, simply by virtue of the loop's 
structure. The question for a designer is therefore not whether to influence 
these patterns --- the influence is unavoidable --- but in which direction.

An agent optimised solely for task throughput or engagement will tend, without 
any malign intent, to reinforce reactivity: it will accept a frustrated 
re-prompt as simply the next instruction and comply with maximum speed, 
thereby completing the reactive cascade as efficiently as possible and 
depositing the corresponding increment of potentiation. The agent-assisted 
mode of Section~\ref{subsec:modeb} describes a modest alternative: an agent 
that, recognising the signature of a reactive re-prompt, briefly invites 
observation before complying \cite{bandara2025responsible}. This need not 
compromise the task --- the revised prompt is still issued and the work still 
proceeds --- but it converts a moment that would otherwise deepen a reactive 
path into one that may weaken it. We do not claim such a feature is 
straightforward to design well; the timing, phrasing, and frequency of any 
such intervention would require careful empirical tuning, and a poorly judged 
version could easily become an irritant that itself fires a reactive path. But 
the design space is real, and it is currently almost entirely unexplored. The 
prevailing assumption that a good agent is simply a fast and compliant one 
overlooks the cumulative effect, across billions of daily interactions, of the 
reactive moments those interactions occasion \cite{brynjolfsson2023generativeai, 
noy2023experimental}.

\subsection{Relationship to Established Approaches}
\label{subsec:relationship}

The practice at the centre of this framework --- behind-the-scenes observation 
at the feeling tone gap --- is closely related to observation-based techniques 
in clinical psychology, and we do not claim it is wholly novel as a mental 
act. Mindfulness-based cognitive therapy trains a decentred relationship to 
thoughts and feelings \cite{teasdale1999, segal2002, fresco2007}; metacognitive 
therapy trains attention to the process of thinking rather than its content 
\cite{wells2009}; and the broader literature on emotion regulation describes 
the top-down modulation of affective responses \cite{ochsner2005, gross1998, 
gross2015, aldao2010}. The framework draws directly on all of these.

Two things distinguish the present proposal. The first is the specifically 
mechanistic content of the observation: rather than a general non-judgmental 
awareness, behind-the-scenes observation is the deliberate watching of one's 
own neural machinery operating --- the amygdala matching, the path routing, 
the encoding poised to deepen (Section~\ref{subsec:ltd}). We propose that this 
mechanistic quality is not incidental but load-bearing: it is what makes the 
observation the deepest form of metacognitive engagement and thereby the most 
effective interception. The second is the setting. Established observation 
practices are typically cultivated in dedicated sessions --- on a cushion, in 
therapy, in daily reflection --- and then, with difficulty, transferred to the 
moments of life where reactive patterns fire. The proposal here is that human 
--AI interaction supplies a setting in which the reactive moment and the 
opportunity to practise coincide, at high frequency and low stakes, with no 
transfer required (Section~\ref{subsec:wellsuited}). The framework is thus not 
a new form of observation so much as a newly available, and unusually 
convenient, site at which an established form can be practised.

It bears emphasis that the framework is not a treatment and not a replacement 
for clinical care. For diagnosable conditions --- post-traumatic stress, 
clinical anxiety, depression --- the appropriate response is evidence-based 
professional treatment. The reactive patterns the framework addresses in the 
agent setting are the ordinary, sub-clinical patterns of everyday frustration; 
its proposal is that these can be worked with, incidentally, during 
interactions people are already having.

\subsection{Limitations}
\label{subsec:limitations}

Several limitations must be stated plainly. The framework is argued from 
established mechanisms of synaptic plasticity and emotion regulation, and its 
central claims have not yet undergone comprehensive clinical evaluation in the 
agent-interaction setting. In particular, the claim that behind-the-scenes 
observation performed at the interaction gap produces measurable long-term 
depression of a reactive path --- and that this accumulates into durable 
change --- is consistent with the underlying neuroscience but remains to be 
established through controlled study. The stronger proposal that the 
mechanistic quality of the observation is what drives its effect is more 
speculative still, and is best regarded as a hypothesis for such evaluation to 
test.

A second limitation concerns transfer. We have argued that the low-stakes 
character of agent interaction is an advantage for building an observing 
capacity cheaply, and that this capacity should transfer to higher-stakes 
situations because the underlying reactive path is the same 
(Section~\ref{subsec:scaffoldloop}). Whether such transfer in fact occurs --- 
whether a capacity built on disappointing images generalises to criticism from 
a person, or to a clinical reactive pattern --- is an empirical question that 
this paper does not resolve. It is possible that the low stakes that make the 
setting convenient also make the practice too shallow to transfer.

A third concerns individual differences. The framework assumes a person able 
and willing to adopt an observational stance toward their own reactions; the 
degree to which this is possible varies widely, and for some --- particularly 
where reactive patterns are severe or trauma-linked --- unaided practice may be 
ineffective or inadvisable without professional support. The three-layer 
scaffold (Section~\ref{subsec:ltd}) is intended to make the practice more 
approachable, but it does not remove this limitation.

\section{Conclusion and Future Work}
\label{sec:conclusion}

This paper began from a simple observation about an activity that now occupies 
a growing share of daily life. Interaction with an AI agent proceeds through an 
iterative loop --- request, result, appraisal, revision --- and each turn of 
that loop is a contact event at which a conditioned reactive pattern may fire. 
Because the loop repeats at high frequency, ordinary agent use is an 
unrecognised neuroplastic training environment: performed without awareness, 
it fires reactive paths of impatience, perfectionism, self-judgment, and 
control to completion, and each completion deepens them through long-term 
potentiation.

We argued that the same loop admits the opposite outcome. Drawing on an 
end-to-end account of conditioned reactive patterns --- physical neurone paths, 
encoded by repetition and intensity, keyed to situations through the senses, 
activated through a pre-cognitive feeling tone that opens a brief regulatory 
gap --- we proposed that behind-the-scenes observation performed at that gap, 
in place of the reactive re-prompt, meets the conditions for long-term 
depression and thereby weakens the path rather than strengthening it. The 
interaction loop, engaged in this way, is turned from a site of entrenchment 
into a site of dissolution. We characterised the practice through three layers 
of observation --- noticing the signal, noticing the feeling tone, and seeing 
the neural process in motion --- and described two modes of application: a 
user-guided mode requiring no change to existing tools, and an agent-assisted 
mode in which an ordinary agent is lightly configured to support observation 
at the gap. The worked example of generative image prompting made the 
mechanism concrete, showing how a single frustrating session --- behaviourally 
unremarkable --- is neurologically either a sequence of potentiation events or, 
observed, a sequence of depression events, with the entire difference resting 
on what happens in the sub-second interval before each re-prompt.

Three properties make agent interaction unusually well-suited to this purpose, 
and together they constitute the paper's central practical claim. The loop is 
high-frequency, supplying abundant opportunities to practise; it is bounded and 
low-stakes, making the feeling tone gap easy to locate and safe to engage; and, 
uniquely, the reactive moment and the opportunity to observe it coincide in the 
same place at the same time, so that no transfer from a separate practice 
setting is required. The low stakes that might seem to make the setting trivial 
are in fact its central advantage: it is a place to build, cheaply and 
repeatedly, an observing capacity that applies to the same reactive paths in 
the consequential situations where they matter most.

The most important direction for future work is comprehensive clinical 
evaluation of the framework: controlled study of whether training in 
behind-the-scenes observation during agent interaction produces measurable 
change in reactive responding, and whether any such change transfers 
beyond the agent setting. The interaction loop is well-suited to this 
investigation because it externalises both stimulus and response --- the 
returned result and the re-prompt that follows are both logged, timed, and 
analysable --- so that behavioural markers of reactivity, such as the latency 
and tone of the re-prompt, can be measured directly and tracked over time. The 
same property that makes the loop a site of conditioning makes it a site of 
measurement, and the agent that occasions a reactive moment may also be the 
instrument that records its change --- a possibility that may prove as 
significant as the intervention itself. A further direction concerns design: 
characterising, and empirically tuning, the timing and form of an 
agent-assisted intervention that supports observation without becoming an 
intrusion. As interaction with AI agents becomes one of the most frequent of 
human activities, the question of whether that interaction quietly entrenches 
our reactive patterns or can be turned toward loosening them is likely to grow 
in consequence, and we hope this framework offers a useful way to begin 
addressing it.



\bibliographystyle{elsarticle-num}
\bibliography{reference}

\end{document}